\documentclass[journal]{IEEEtran}
\usepackage[pdftex]{graphicx}
\usepackage{cite}
\usepackage{amsmath}
\usepackage{multirow}
\usepackage{color}
\usepackage[switch]{lineno}
\usepackage{graphicx}
\usepackage{subfigure}
\usepackage{url}
\usepackage{epstopdf}
\usepackage{hyperref}

\usepackage{lineno}
\usepackage{ulem}
\usepackage{verbatim}

\ifCLASSINFOpdf
\else
\fi

\begin{document}

\title{Facial Landmark Machines: A Backbone-Branches Architecture with Progressive Representation Learning}
\author{Lingbo~Liu,
        Guanbin~Li,
        Yuan~Xie,
        Yizhou~Yu,
        Qing~Wang
        and~Liang~Lin
\thanks{Corresponding author is Guanbin Li~(e-mail:liguanbin@mail.sysu.edu.cn). }
\thanks{L. Liu, G. Li, Y. Xie, Q.Wang and L. Lin are with the School of Data and Computer Science, Sun Yat-sen University, Guangzhou 510006, China. Y. Yu is with Department of Computer Science, The University of Hong Kong, Hong Kong.}

\thanks{Project website:  \color{blue}\href{url}{http://www.sysu-hcp.net/facial-landmark-localization/}}
}
\maketitle

\begin{abstract}
Facial landmark localization plays a critical role in face recognition and analysis. In this paper, we propose a novel cascaded backbone-branches fully convolutional neural network~(BB-FCN) for rapidly and accurately localizing facial landmarks in unconstrained and cluttered settings. Our proposed BB-FCN generates facial landmark response maps directly from raw images without any preprocessing. BB-FCN follows a coarse-to-fine cascaded pipeline, which consists of a backbone network for roughly detecting the locations of all facial landmarks and one branch network for each type of detected landmark for further refining their locations.
Furthermore, to facilitate the facial landmark localization under unconstrained settings, we propose a large-scale benchmark named SYSU16K, which contains 16000 faces with large variations in pose, expression, illumination and resolution.
Extensive experimental evaluations demonstrate that our proposed BB-FCN can significantly outperform the state-of-the-art under both constrained (i.e., within detected facial regions only) and unconstrained settings. We further confirm that high-quality facial landmarks localized with our proposed network can also improve the precision and recall of face detection.
\end{abstract}

\IEEEpeerreviewmaketitle

\section{Introduction}
\IEEEPARstart{F}acial landmark localization aims to automatically predict key point positions in facial image regions. This task is an essential component in many face-related applications, such as facial attribute analysis~\cite{luo2013deep}, face verification~\cite{lu2015surpassing,liu2016deep} and face recognition~\cite{zhu2013deep,ding2015robust,li2017face}. Although tremendous effort has been devoted to this topic, its performance is still far from perfect, particularly on facial regions with severe occlusions or extreme head poses.

\begin{figure}
\begin{center}
   \includegraphics[width=0.95\columnwidth]{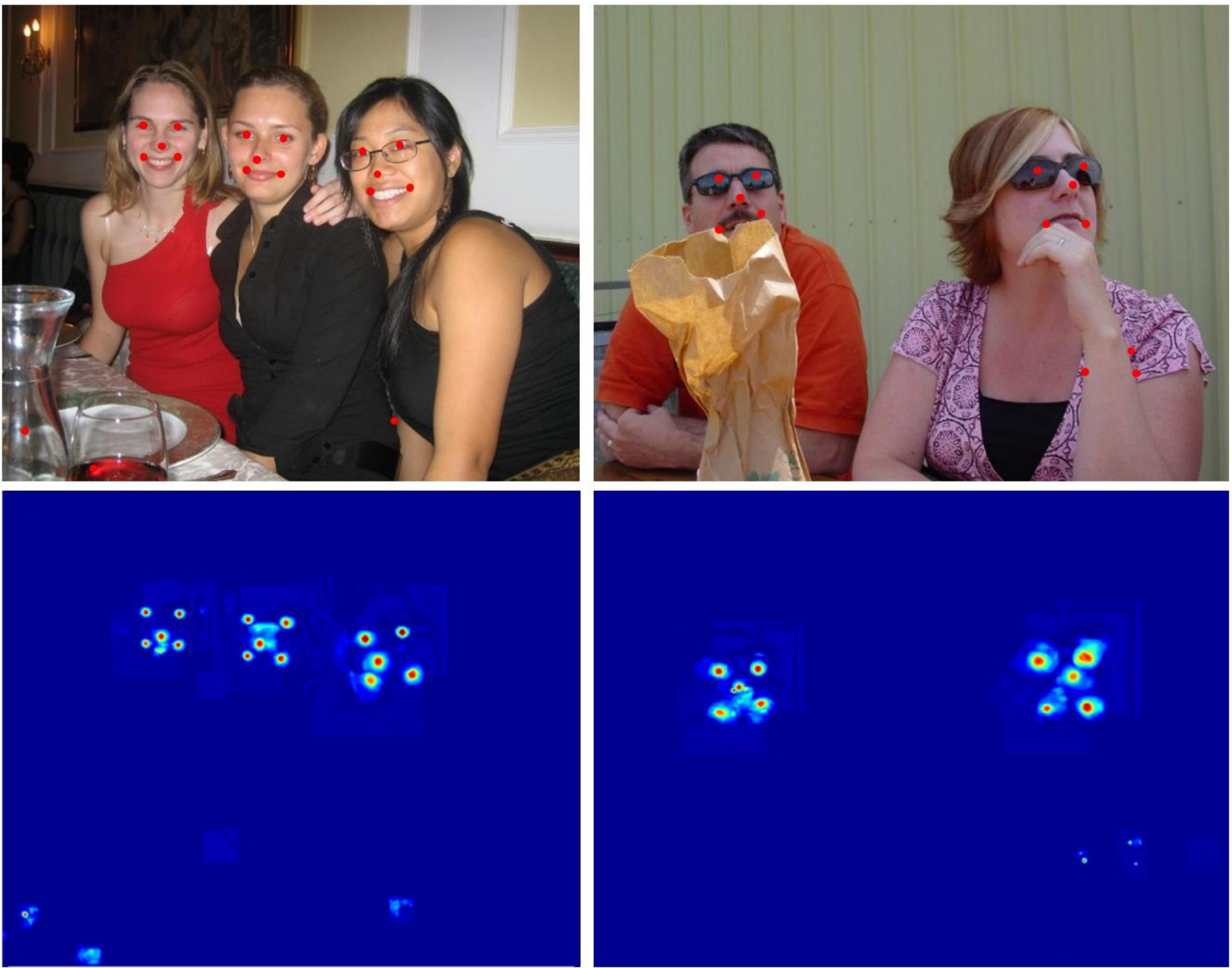}
\end{center}
\vspace{-4mm}
\caption{Facial landmark localization in an unconstrained setting. (a) Two cluttered images with an unknown number of faces. (b) Dense response maps generated by our method.}
\vspace{-3mm}
\label{fig:unconstrained-example}
\end{figure}

Most of the existing approaches for facial landmark localization have been developed for a controlled setting, e.g., the facial regions are detected in a preprocessing step. This setting has drawbacks when we work with images taken in the wild (e.g., cluttered surveillance scenes), where automated face detection is not always reliable. The objective of this work is to propose an effective and efficient facial landmark localization method that is capable of handling images taken in unconstrained settings and that contain multiple faces, extreme head poses and occlusions~(see Figure~\ref{fig:unconstrained-example}).  Specifically, we keep the following issues in mind when developing our algorithm.

\begin{itemize}
\item Faces may have large appearance and structure variations in  unconstrained settings due to diverse viewing conditions, rich facial expressions, large pose changes, facial accessories (e.g., glasses and hats) and aging. Therefore, traditional global models may not work well because the usual assumptions (e.g., certain spatial layouts) may not hold in such environments.
\item Boosted-cascade-based fast face detectors, which evolved from the seminal work of Viola and Jones~\cite{viola2001rapid}, can only work well for near-frontal faces under normal conditions. Although accurate deformable-part-based models~\cite{zhu2012face} can perform much better on challenging datasets, these models are slow due to their high complexity. Detection in an image takes a few seconds, which makes such detectors impractical for our task.
\end{itemize}

In this paper, we formulate facial landmark localization as a pixel-labeling problem and develop a fully convolutional neural network (FCN) to overcome the aforementioned issues. The proposed approach produces facial landmark response maps directly from raw images without relying on any preprocessing or feature engineering. Two typical landmark response maps generated with our method are shown in Figure~\ref{fig:unconstrained-example}.

With the recent advances in deep learning techniques and large-scale annotated image datasets, such as ImageNet, deep convolutional neural network models have achieved significant progress in generic object detection~\cite{girshick2014rich}, crowd analysis~\cite{liu2018crowd,liu2018attentive} and facial landmark localization~\cite{zhang2014facial}. Facial landmark localization is typically formulated as a regression problem. Among the existing methods that take this approach, the cascaded deep convolutional neural networks~\cite{sun2013deep,weng2016learning} have emerged as one of the leading methods because of their superior accuracy. Nevertheless, this three-level cascaded CNN framework is complicated and unwieldy. It is arduous to jointly handle the classification (i.e., whether a landmark exists) and localization problems for unconstrained settings. Long et al.~\cite{long2015fully} recently proposed an FCN for pixel labeling, which takes an input image with an arbitrary size and produces a dense label map in the same resolution. This approach shows convincing results for semantic image segmentation and is also very efficient since convolutions are shared among overlapping image patches. Notably, classification and localization can be simultaneously achieved with a dense label map. The success of this work inspires us to adopt an FCN in our task, i.e., pixelwise facial landmark prediction. Nevertheless, a specialized architecture is required because our task requires more accurate prediction than generic image labeling.

Considering both computational efficiency and localization accuracy, we pose facial landmark localization as a cascaded filtering process. In particular, the locations of facial landmarks are first roughly detected in a global context, and then they are refined by observing local regions. To this end, we introduce a novel FCN architecture that naturally follows this coarse-to-fine pipeline. Specifically, our architecture contains one backbone network and several branches, with each branch corresponding to one landmark type. For computational efficiency, the backbone network is designed to be an FCN with lightweight filters, which takes a low-resolution image as its input and rapidly generates an initial multichannel heat map with each channel predicting the location of a specific landmark. We can obtain landmark proposals from each channel of the initial heat map. We then crop a region centered at every landmark proposal from both the original input image and the corresponding channel of the response map. These cropped regions are stacked together and fed to a branch network for a fine and accurate localization.
Because fully connected layers are not used in either network, we call our architecture the cascaded backbone-branches fully convolutional network~(BB-FCN). Thanks to the tailor-designed architecture of the backbone network, which can reject most background regions and retain high-quality landmark proposals, our BB-FCN is also capable of accurately localizing the landmarks of various scale faces by rapidly scanning every level of the constructed image pyramid.
Furthermore, we have also discovered that our landmark localization results can help generate fewer and higher-quality face proposals,  thus enhancing the accuracy and efficiency of face detection.

In summary, our contributions in this paper can be summarized as follows:
\begin{itemize}
\item We propose a new BB-FCN architecture for facial landmark localization, which consists of a backbone network for rough landmark prediction and a set of branch networks, where each network is for refining the predictions of one specific type of landmark. 
\item We extensively evaluate BB-FCN on several standard benchmarks (e.g., AFW~\cite{zhu2012face}, AFLW~\cite{kostinger2011annotated} and 300W~\cite{sagonas2013300}), and our experiments show that BB-FCN achieves superior performance in comparison to other state-of-the-art methods under both constrained (i.e., with face detections) and unconstrained settings. In particular, our BB-FCN significantly decreases the average mean error of the current best-performing method from 8.2\% to 6.18\% on AFW and from 6.58\% to 6.28\% on AFLW.
\item We use our facial landmark localization results to guide R-CNN-based face detection and demonstrate significant increases in both accuracy and efficiency.
\end{itemize}

The remainder of this paper is organized as follows. Section~\ref{sec:relatedwork} discusses related work and differentiates our method from such works. Section~\ref{sec:bb_fcn} introduces our proposed BB-FCN architecture. The experimental results and comparisons are presented in Section~\ref{sec:experiment}. Finally, Section~\ref{sec:conclusion} concludes this paper.

\section{Related Work}\label{sec:relatedwork}
Facial landmark localization has long been attempted in computer vision, and a large number of approaches have been proposed for this purpose. The conventional approaches for this task can be divided into two categories: template fitting methods and regression-based methods.

Template fitting methods build face templates to fit input face appearance~\cite{cootes2001active}. A representative work is the active appearance model~(AAM)~\cite{cootes2001active}, which attempts to estimate model parameters by minimizing the residual between the holistic appearance and an appearance model. A vast collection of methods based on AAM have been proposed~\cite{sauer2011accurate, tresadern2010additive, tzimiropoulos2013optimization}. Rather than using holistic representations, a constrained local model~(CLM)~\cite{saragih2011deformable} learns an independent local detector for each facial keypoint and a shape model for capturing valid facial deformations. Improved versions of CLM primarily differ from each other in terms of local detectors. For instance, Belhumeur~{\em et al.}~\cite{belhumeur2013localizing} detected facial landmarks by employing SIFT features and SVM classifiers, and Liang~{\em et al.}~\cite{liang2008face} applied AdaBoost to the HAAR wavelet features. 
These methods are generally superior to the holistic methods due to the robustness of patch detectors against illumination variations and occlusions.

Regression-based facial landmark localization methods can be further divided into direct mapping techniques and cascaded regression models. The former directly maps local or global facial appearances to landmark locations. For example, Dantone {\em et al.}~\cite{dantone2012real} estimated the absolute coordinates of facial landmarks directly from an ensemble of conditional regression trees trained on facial appearances. Valstar {\em et al.}~\cite{valstar2010facial} applied boosted regression to map the appearances of local image patches to the positions of corresponding facial landmarks. Cascaded regression models~\cite{cao2014face, kazemi2014one, xiong2013supervised, ren2014face,zhu2016unconstrained,tuzel2016robust,fan2017explicit} formulate shape estimation as a regression problem and make predictions in a cascaded manner. These models typically start from an initial face shape and iteratively refine the shape according to learned regressors, which map local appearance features to incremental shape adjustments until convergence. Cao~{\em et al.}~\cite{cao2014face} trained a cascaded nonlinear regression model to infer an entire facial shape from an input image using pairwise pixel-difference features. Burgos-Artizzu~{\em et al.}~\cite{burgos2013robust} proposed a novel cascaded regression model for estimating both landmark positions and their occlusions using robust shape-indexed features. Another seminal method is the supervised descent method~(SDM)~\cite{xiong2013supervised}, which uses SIFT features extracted around the current shape and minimizes a nonlinear least-squares objective using the learned descent directions. All these methods assume that an initial shape is given in some form, e.g., a mean shape~\cite{ren2014face,xiong2013supervised}. However, this assumption is too strict and may lead to poor performance on faces with large pose variations.

Despite acknowledged successes, all the aforementioned conventional approaches rely on complicated feature engineering and parameter tuning, which consequently limits their performance in cluttered and diverse settings. Recently, convolutional neural networks and other deep learning models have been successfully applied to various visual computing tasks, including facial landmark estimation. Zhou~{\em et al.}~\cite{zhou2013extensive} proposed a four-level cascaded regression model based on CNNs, which sequentially predicted landmark coordinates. Zhang~{\em et al.}~\cite{zhang2014facial} employed a deep architecture to jointly optimize facial landmark positions with other related tasks, such as pose estimation~\cite{liu2016sparse} and facial expression recognition~\cite{zhang2016deep}. Zhang~{\em et al.}~\cite{zhang2014coarse} proposed a new coarse-to-fine DAE pipeline to progressively refine facial landmark locations. In 2016, they further presented  de-corrupt autoencoders to automatically recover the genuine appearance of the occluded facial parts, followed by predicting the occlusive facial landmarks~\cite{zhang2016occlusion}. Lai~{\em et al.}~\cite{lai2015deep} proposed an end-to-end CNN architecture to learn highly discriminative shape-indexed features and then refined the shape using the learned deep features via sequential regressions.
Merget~{\em et al.}~\cite{merget2018robust} integrated the global context in a fully convolutional network based on dilated convolutions for generating robust features for landmark localization.
Bulat~{\em et al.}~\cite{bulat2017super} utilized a facial super-resolution technique to locate the facial landmarks from low-resolution images.
Tang~{\em et al.}~\cite{tang2018quantized} proposed  quantized densely connected U-Nets to largely improve the information flow, which helps to enhance the accuracy of landmark localization.
RNN-based models~\cite{peng2016recurrent,xiao2016robust,trigeorgis2016mnemonic} formulate facial landmark detection as a sequential refinement process in an end-to-end manner.
Recently, 3D face models~\cite{zhu2016face, jourabloo2016large, liu2016joint,bulat2017far,feng2018joint} have also been utilized to accurately locate the landmarks by modeling the structure of facial landmarks. Moreover, many researchers have attempted to adapt some unsupervised \cite{dong2018supervision,zhang2018unsupervised,dong2018style} or semisupervised \cite{honari2018improving} approaches to improve the precision of facial landmark detectors.

Although these methods have achieved remarkable performance, most of them were developed for a controlled setting, which requires a detected frontal face as the input. These methods basically pose landmark estimation as a parameterized regression process, e.g., mapping landmark coordinates, which actually restricts the flexibility in practice due to the fixed form of the parameterization.  Such trained models struggle in unconstrained settings (e.g.,  unknown number of faces in an image). In contrast, our approach produces pixelwise response maps, making it very flexible in localizing facial landmarks in the wild and in integrating with other methods. 

\begin{figure*}
\begin{center}
 \includegraphics[width=1.9\columnwidth]{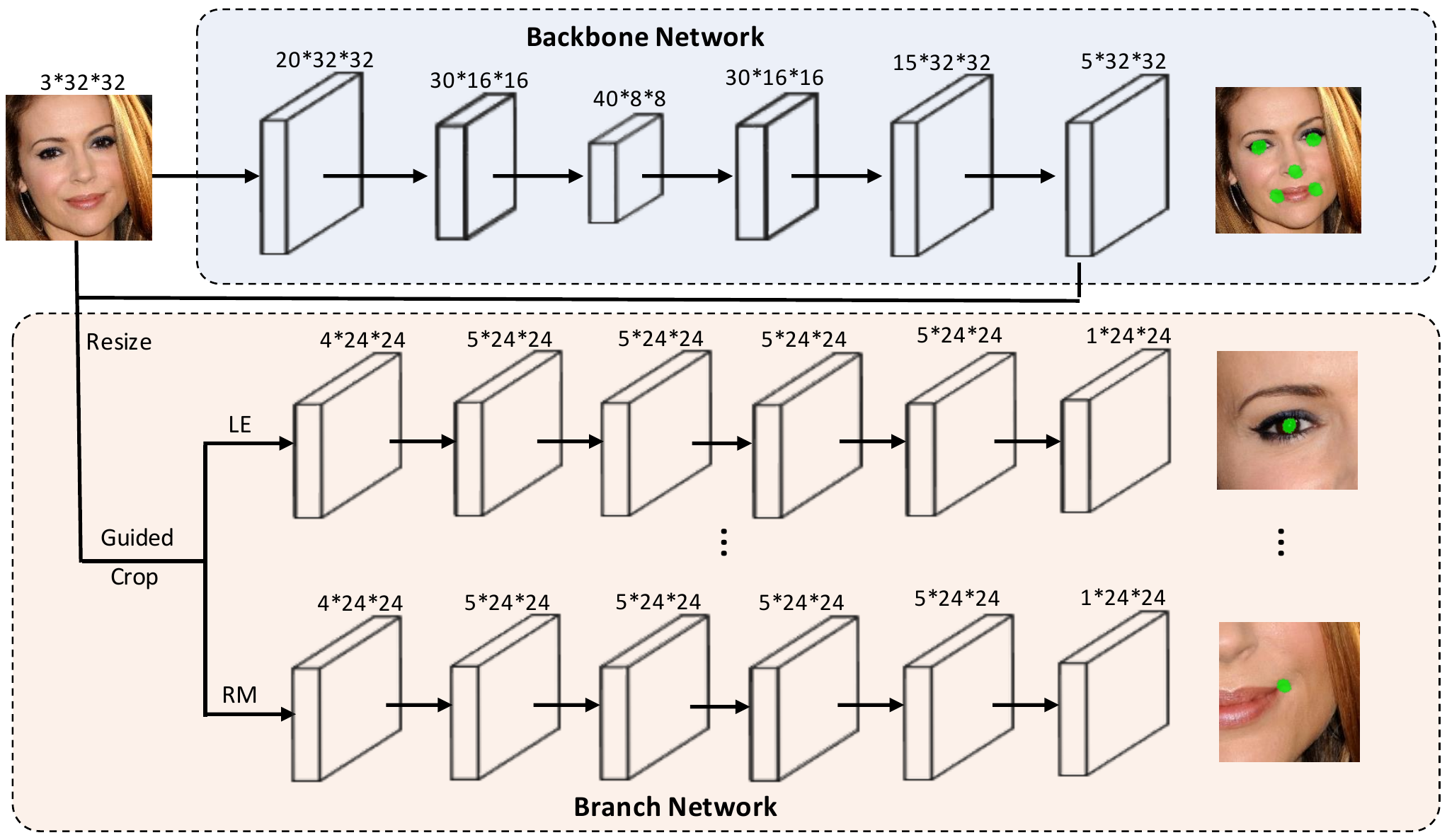}
 \vspace{-4mm}
\end{center}
   \caption{The main architecture of the proposed backbone-branches fully convolutional neural network. This approach is capable of producing pixelwise facial landmark response maps in a progressive manner. The backbone network first generates low-resolution response maps that identify approximate landmark locations via a fully convolutional network. The branch networks then produce fine response maps over local regions for more accurate landmark localization. There are $K$~(e.g., $K=5$) branches, each of which corresponds to one type of facial landmark and refines the related response map. Only downsampling, upsampling, and prediction layers are shown, and intermediate convolutional layers are omitted in the network branches.}
\vspace{-4mm}
\label{fig:network-structure}
\end{figure*}

\section{The Cascaded BB-FCN Architecture}\label{sec:bb_fcn}
Given an unconstrained image $I$ with an unknown number of faces, our facial landmark localization method aims to locate all facial landmarks in the image. We use $L_i^k=(x_i^k,y_i^k)$ to denote the location of the $i^{th}$ landmark of type $k$ in image $I$, where $x_i^k$ and $y_i^k$ represent the coordinates of this landmark. Then, our task is to obtain the complete set of landmarks in $I$,
\begin{equation}
Det(I)=\{(x_i^k,y_i^k)\}_{i,k},
\end{equation}
where $k=1,2,...,K$. When describing our method and analyzing the proposed network, we set $K=5$ as an example, but our method is also applicable to any other values of $K$. In the experimental section, we will also present simultaneous localization results for 29 landmark types and 68 landmark types. Here, the five landmark types are the left eye (LE), right eye (RE), nose (N), left mouth corner (LM) and right mouth corner (RM).

In contrast to existing approaches that predict landmark locations through coordinate regression, we exploit FCNs to directly produce response maps that indicate the probability of landmark existence at every image location. FCNs have shown excellent performance in various pixel-labeling problems, such as semantic image segmentation~\cite{long2015fully}, object contour detection~\cite{xie2015holistically} and salient object detection~\cite{li2016visual,chen2016disc,li2018contrast}. Applying an FCN to an image resembles a deep filtering process. An FCN naturally operates on an input image of any size, and it produces an output with the corresponding spatial dimensions. In our method, the predicted value at each location of the response map can be viewed as a series of filtering operations applied to a specific region of the input image. This specific region is called the receptive field. An ideal series of filters should have the following property: a receptive field with a landmark of a specific type located at its center should return a strong response value, whereas receptive fields without that type of landmark in the center should yield weak responses. Let $F_{\mathbf W^k}(P)$ denote the result of applying a series of filtering functions with parameter setting $\mathbf W^k$ for type-$k$ landmarks to receptive field $P$, and it is defined as follows:
\begin{equation}
\label{equ:response-func}
F _{\mathbf W^k}(P) =
\begin{cases}
1  & \text{if $P$ has a type-$k$ landmark in the center};\\
0 & \text{otherwise}.
\end{cases}
\end{equation}
Applying this function in a sliding window manner to $w \times h$ overlapping receptive fields in an input image $I$ generates a response map $F_{\mathbf W^k}*I$ of size $w \times h$, whose value at location $(x,y)$ can be defined as
\begin{equation}
\label{equ:reponse}
(F_{\mathbf W^k}*I)(x,y)=F_{\mathbf W^k}(I(P(x,y))),
\end{equation}
where $I(P(x,y))$ stands for the image patch corresponding to the receptive field of location~(x,y) in the output response map.

If the response value is larger than a threshold $\theta$, a landmark of type $k$ is detected at the center of the patch in image $I$. Thus,
\begin{equation}
Det(I)\!\!=\!\!\{(\text{center of }P(x,y))|(F_{\mathbf W^k}*I)(x,y)>\theta \}.
\end{equation}

According to Equation (\ref{equ:reponse}), there is a trade-off between localization accuracy and computational cost.  To achieve high accuracy, we need to compute response values for significantly overlapping receptive fields. However,  to accelerate the detection process, we should generate a coarse response map on less overlapping receptive fields or from a lower-resolution image. This motivates us to develop a cascaded coarse-to-fine process to localize landmarks progressively, in a spirit similar to the hierarchical deep networks in \cite{HDCNNyan2015} for image classification.  Specifically, the architecture of our deep network consists of two components. The first component generates a coarse response map from a relatively low-resolution input, identifying rough landmark locations. Then, the other component takes local patches centered at every estimated landmark location and applies another filtering process to the local patches to obtain a fine response map for accurate landmark localization. This cascaded two-stage strategy enables us to accurately detect facial landmarks at a high speed.

In this paper, this two-component architecture is implemented as a BB-FCN, where the backbone network generates coarse response maps for rough location inference and the branch networks produce fine response maps for accurate location refinement. Figure~\ref{fig:network-structure} shows the architecture of our network.

Let a convolutional layer be denoted as $C(n, h \times w \times ch)$ and a deconvolutional layer be denoted as $D(n, h \times w \times ch)$, where $n$ represents the number of kernels and $h$, $w$, and $ch$ respectively represent the height, width and number of channels of a kernel. We also use $MP$ to denote a max-pooling layer. In our backbone-branch network, the stride of all convolutional layers is $1$, and the stride of all deconvolutional layers is $2$. The size of the max-pooling operator is set to $2 \times 2$, and the stride for pooling is $2$.

\subsection{Backbone Network}
The backbone network is an FCN. It can efficiently generate an initial low-resolution response map for input image $I$. When localizing facial landmarks in an image taken in an unconstrained setting, it can effectively reject a majority of background regions with a threshold. Let $\mathbf W_c$ denote its parameters and $H^k(I;\mathbf W_c)$ denote the predicted heat map of image $I$ for the $k^{th}$ type of landmarks. The value of $H^k(I;\mathbf W_c)$ at position $(x,y)$ can be computed with Equation (\ref{equ:reponse}).
We train the backbone FCN using the following loss function:
\begin{equation}
\label{equ:backbone_loss}
\mathcal L_{1}(I;\mathbf W_c)=\sum_{k=1}^K{||H^k(I;\mathbf W_c)-H^k_c(I)||^2},
\end{equation}
where $H^k_c(I)$ denotes the ground-truth heat map for type-$k$ landmarks.

The backbone network is trained with a patch-based optimization scheme. During the training phase, the human faces are cropped from the unconstrained crowded images and resized to a low resolution of $32 \times 32$. Taking the cropped patches of whole faces as input, the backbone network can implicitly learn the geometric constraints among landmarks and generate the response heat maps of all facial landmarks together.
Specifically, the backbone network consists of eight convolutional layers with lightweight filters and two deconvolutional layers, which are detailed as follows:
$C(20, 5 \times 5 \times 3 )$ - $C(20, 5 \times 5 \times 20 )$ - $MP$ -
$C(30, 5 \times 5 \times 20 )$ - $C(30, 5 \times 5 \times 30 )$ - $MP$ -
$C(40, 5 \times 5 \times 30 )$ - $C(40, 5 \times 5 \times 40 )$ -
$D(30, 2 \times 2 \times 40 )$ - $C(30, 5 \times 5 \times 30 )$ -
$D(15, 2 \times 2 \times 30)$ - $C(5, 1 \times 1 \times 15)$.

\subsection{Branch Network}
The branch network is composed of $K$ branches, with each branch responsible for detecting one type of landmark. All the $K$ branches are designed to share the same network structure. In branch networks, a cropped patch from the original input image and a region from the backbone's output heat map are stacked together as its input. Therefore, the input data consist of four channels, including $3$ channels from the original $RGB$ image and $1$ channel from the corresponding channel of the backbone's output heat map.  To make the branch network better suited for landmark position refinement, we resize the original input image to $64 \times 64$, which is four times the size of the backbone's input, and simultaneously magnify the heat map from the backbone network to $64 \times 64$. The resolution of all the cropped patches is $24 \times 24$, and they are all centered at the landmark position predicted by the backbone network. As shown in Fig.~\ref{fig:network-structure}, each branch is trained in the same way as the backbone network. We denote the parameters of the branch component for type-$k$ landmarks as $\mathbf W_f^k$, and we respectively use $H(P;\mathbf W_f^k), H_0^k(P)$ to denote the predicted fine heat map and the corresponding ground-truth heat map of patch $P$. The loss function of this branch component is again defined as follows:
\begin{equation}
\label{equ:branch_loss}
\mathcal L_2(P;\mathbf W_f^k)=||H(P;\mathbf W_f^k)-H_0^k(P)||^2.
\end{equation}
Each branch component is composed of $5$ convolutional layers without any pooling operations. The dimensionality of its input data is $24 \times 24 \times 4$. The first $4$ convolutional layers consist of $5$ channels with the kernel size equal to $5$ and stride equal to $1$, while the last convolutional layer consists of $5$ channels with a kernel size of $1$ and stride of $1$. As shown in Figure~\ref{fig:network-structure}, each branch FCN component is detailed as follows:
$C(5, 5 \times 5 \times 4 )$ - $C(5, 5 \times 5 \times 5 )$ -
$C(5, 5 \times 5 \times 5 )$ - $C(5, 5 \times 5 \times 5 )$ - $C(1, 1 \times 1 \times 5 )$.

\begin{figure}[t]
\centering
   \includegraphics[width=0.85\columnwidth]{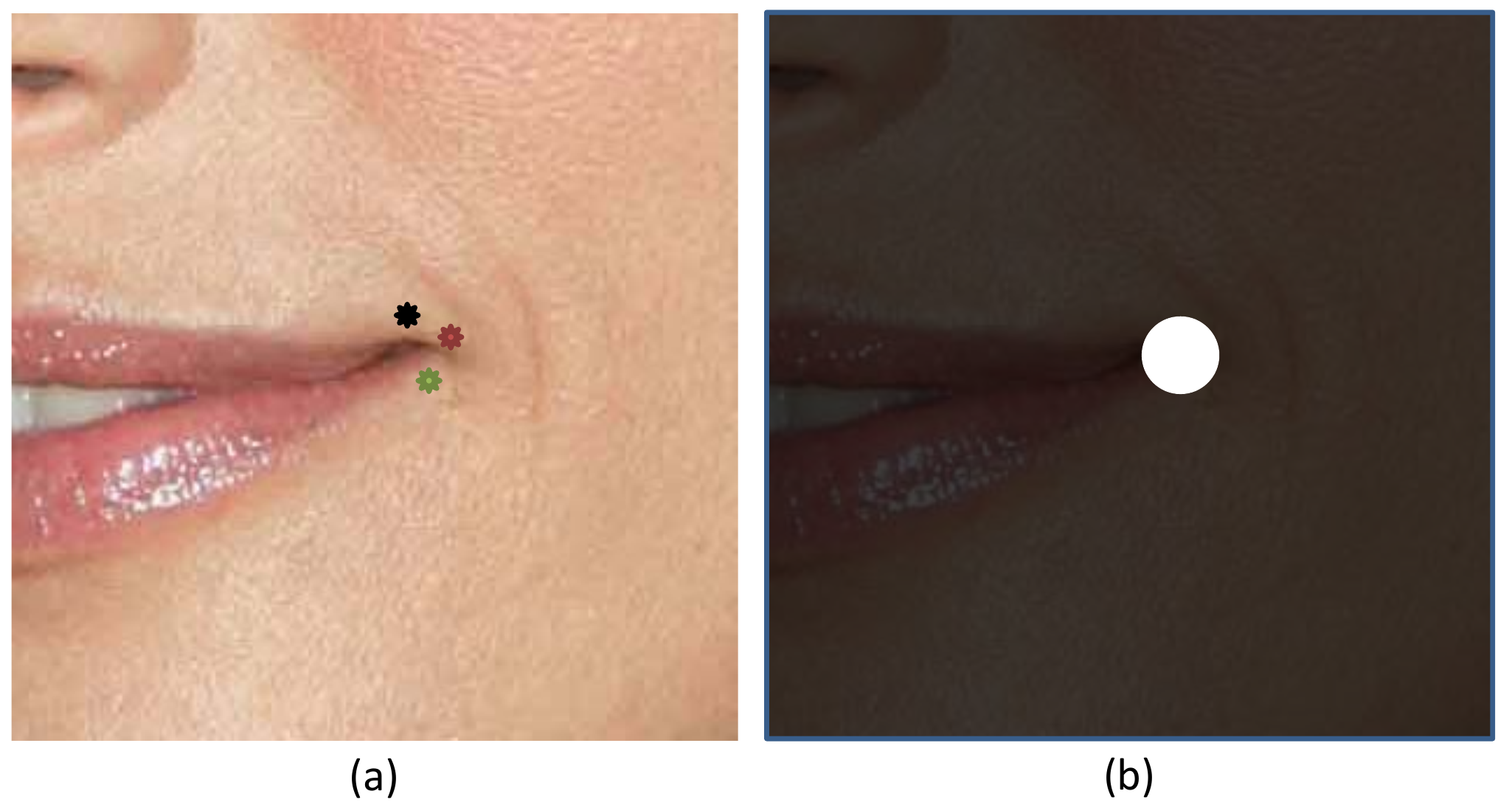}
\vspace{-4mm}
\caption{(a) An isolated point cannot accurately reflect discrepancies among multiple annotations. The three points near the right mouth corner were annotated by three different workers. (b) We label a landmark as a small circular region rather than an as an isolated point in the ground-truth heat map.}
\vspace{-4mm}
\label{fig:gt_generation}
\end{figure}

\subsection{Ground-truth Heat Map Generation}
To our knowledge, the ground truth of a facial landmark is traditionally given as a single pixel location~$(x,y)$ in all public datasets. To adapt such landmark specifications for the training stage of our proposed BB-FCN network, we generate the ground-truth heat map of an input image according to the annotated facial landmark locations. The most straightforward method assigns ``$1$'' to a single pixel corresponding to each landmark location and ``$0$'' to the remaining pixels.
However, we argue that this method is suboptimal because an isolated point cannot reflect discrepancies among multiple annotations. As shown in Figure~\ref{fig:gt_generation}(a), the right mouth corner has three slightly different locations marked by three annotators. To take such discrepancies into consideration, we label each landmark as a small region rather than as an isolated point. We first initialize the heat map with zeros everywhere, and then for each landmark $p$, we mark a circular region with center $p$ and radius $R$ in the ground-truth heat map with $1$. 
Different radii are adopted for the backbone network and branch networks, denoted as $R_c$ and $R_f$, respectively. $R_f$ is set to be smaller than $R_c$ because the backbone network estimates coarse landmark positions while the branch networks predict accurate landmark locations.

\subsection{Selective Response Map Training}\label{sec:selectiveresponsemap}
According to Equations (\ref{equ:backbone_loss}) and (\ref{equ:branch_loss}), the loss is computed over the full response map. However, this approach gives rise to a severe imbalance between positive and negative training samples because landmarks are very sparse. This unbalanced setting could mislead the response map to take all zero values when the loss is minimized. Therefore, we adopt a selective scheme, i.e., randomly choosing the same number of non-landmark locations as landmark locations in the ground-truth response map to propagate the errors while inhibiting all other non-landmark locations during error backpropagation. For some invisible landmarks or background images, the ground-truth maps have no positive region, and we only select a small ratio of the non-landmark locations to propagate. This selective training scheme is critical in ensuring the convergence of training sessions in our experiments. In addition, for more effective training and more precise results, hard negative mining is also employed. In the selective phase, hard negative samples, which are non-landmark locations with large output values, are selected to propagate the errors when the loss on the validation set stops decreasing.

\begin{figure*}[t]
\centering
   \includegraphics[width=2\columnwidth]{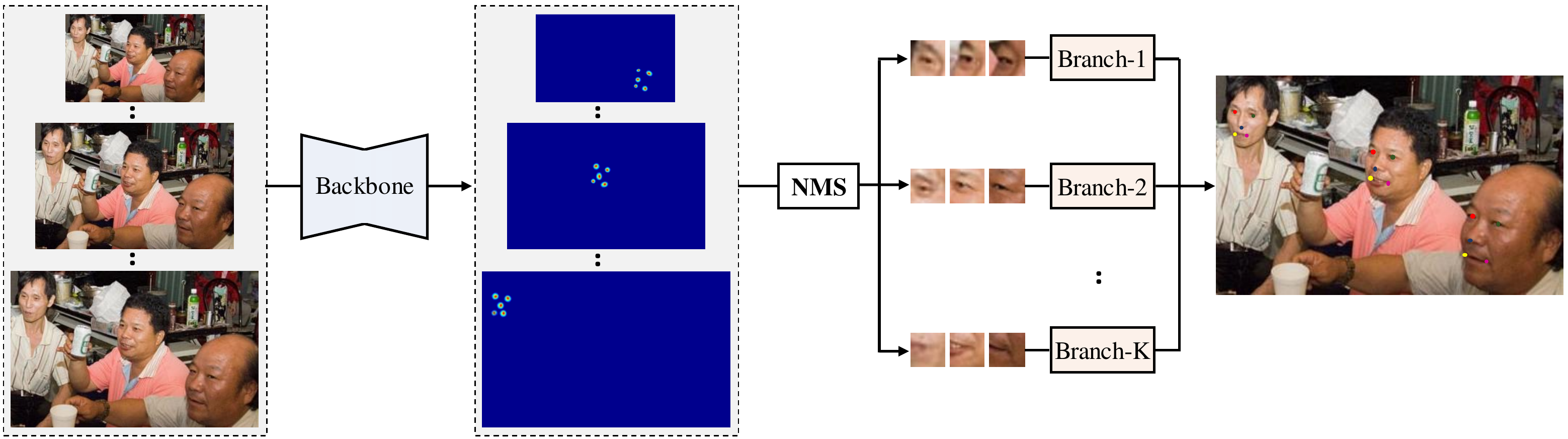}
\vspace{-3mm}
\caption{Illustration of the facial landmark testing procedure under an unconstrained setting. Given an unconstrained image, we first construct an image pyramid. Then, we feed the images at different levels of the pyramid to the backbone network for generating the landmark candidate regions. After adopting a nonmaximum suppression (NMS) to reduce the highly overlapping regions, we refine the locations of the remaining candidate regions with the branch networks. Best viewed in color with magnification.}
\vspace{0mm}
\label{fig:gt_generation}
\end{figure*}

\subsection{Implementation Details}
We have implemented our proposed BB-FCN network in Caffe. A GTX Titan X GPU is used for both training and testing. During training, we randomly initialize our networks by drawing weights from a zero-mean Gaussian distribution with a standard deviation equal to 0.01. The size of a minibatch is set to $40$, and the ratio between the numbers of positive and negative training images in each batch is $1:1$ for the backbone network and $4:1$ for the branch networks. The positive training images are image regions cropped from face images in our SYSU16K dataset, which will be described in Section~\ref{sec:datasets}. The intersection-over-union (IoU) between any cropped region and the original face image is above 0.5. The negative training samples are nonfacial regions randomly cropped from the Pascal VOC 2012 dataset~\cite{everingham2015pascal}. Both the backbone and branch networks are trained using backpropagation and stochastic gradient descent (SGD) with the momentum set to 0.9 and weight decay set to 0.0005. When training the backbone network, we set the learning rate to 0.001 and the total number of iterations to 25K. The radius $R_c$ of landmark circles is set to $5\%$ of the width of the input image. For the branch networks, the total number of iterations is set to 50K. The learning rate is set to $10^{-4}$ for the first 30K iterations and $10^{-5}$ for the last 20K iterations. The radius $R_f$ of landmark circles is set to $3\%$ of the width of the input image. During training, only a subset of the non-landmark locations in the heat map are chosen to propagate errors, as described in Section~\ref{sec:selectiveresponsemap}.

During the testing phase, our BB-FCN network is able to accurately locate facial landmarks under both constrained and unconstrained settings.
For convenience in the following part, we denote the average position of the $n$ locations with the highest response values in a 2D heat map ${M}$ as $Ave\{M, n\}$.

\subsubsection{\textbf{Constrained setting}}
Given a cropped facial image $I$, we first resize it to $32\times32$ and feed it to the backbone network to generate the coarse response heat map $M_c$. Because the radius {$R_c$} is set to ${32 \times {5\%} \approx 2}$, there are 13 pixels in the ground-truth landmark circle of the backbone network. For landmark type $k$, we take $Ave\{M_c^k, 13\}$ as its coarse landmark location, where ${M_c^k}$ is the $k^{th}$ channel of $M_c$.

We resize $I$ and $M_c$ to $64\times64$. For landmark type $k$, we crop a $24\times24$ patch centered at the coarse landmark location from the concatenation of $I$ and $M_c^k$, and we feed the patch into the $k^{th}$ subnet of the branch networks to generate the fine map $M_f^k$. As the radius {$R_f$} is set to ${64 \times {3\%} \approx 2}$, we take $Ave\{M_f^k, 13\}$ as the final location of landmark type $k$.

\subsubsection{\textbf{Unconstrained setting}}
Given an unconstrained image, we construct an image pyramid of $L$ levels by first resizing the image to make the length of the smaller side equal to $32$ and gradually upsampling it with a scale factor of $1.16$. The level number $L$ can be dynamically adjusted based on the acceptable minimum face size. For example, we set $L$ as 20 to locate the landmarks of the tiny faces in the AFW~\cite{zhu2012face} dataset.

We further feed the images at different pyramid levels to the backbone network for generating multiple coarse heat maps and denote the $k^{th}$ channel of the coarse heat maps at the ${l^{th}}$ level as ${M_{c,l}^k}$. When the response value at location (x,y) of ${M_{c,l}^k}$ is higher than a given threshold, we assert that there is a ${12 \times 12}$ candidate region of landmark type $k$ centered at that position. We denote this candidate region with a tuple $\{k,l,v,(x,y)\}$, where $v$ is the response value at location (x,y).

A single landmark may be detected multiple times at a specific level or at different levels of the image pyramid.  To reduce redundancy, for each landmark type, we first map all landmark candidate regions to the original image and then adopt nonmaximum suppression (NMS) with an IOU threshold of 0.5 on these regions based on their response values. For a remaining landmark candidate region $\{k,l,v,(x,y)\}$, we crop its corresponding ${12 \times 12}$ heat map patch from ${M_{c,l}^k}$ and the RGB patch from the image at the ${l^{th}}$ level of the pyramid and further resize these two patches to $24 \times 24$ before feeding them into the $k^{th}$ subnet of the branch networks to generate the fine heat map $M_f^k$. The final landmark location is computed by $Ave\{M_f^k, 13\}$.

\section {Experimental Results}\label{sec:experiment}
\subsection{Datasets}\label{sec:datasets}
The existing public datasets of facial landmark localization are either too small and contain only hundreds of images or have very limited variation across different samples, e.g., most of the samples are near-frontal faces.
 These two situations greatly limit the performance of facial landmark localization under unconstrained settings.
Therefore, we build a large-scale dataset called SYSU16K, which contains 7317 images (6317 for training and 1000 for validation) with 16K faces collected from the Internet. Each face is accurately annotated with 72 landmarks. With a large variation, the faces in our dataset exhibit various poses, expressions, illuminations and resolutions, and they may have severe occlusions.
In addition, to train our proposed BB-FCN, we also randomly select 7542 natural images (6542 for training and 1000 for validation) without any faces from Pascal-VOC2012 as negative samples.

In our experiment, we  evaluate our method on four public challenging datasets: LFPW~\cite{belhumeur2013localizing}, AFW~\cite{zhu2012face}, AFLW~\cite{kostinger2011annotated} and 300W~\cite{sagonas2013300}. There is no overlap among the training, validation and evaluation datasets.

{\flushleft \textbf{AFLW:}} This dataset contains 21,080 faces in the wild. This dataset is very suitable for evaluating the performance of face alignment across a large range of poses. The selection of testing images from AFLW is as in~\cite{zhang2014facial}, which randomly chooses 3000 faces, and 39\% of them are non-frontal.
{\flushleft \textbf{AFW:}} This dataset contains 205 images (468 faces) collected in the wild. Invisible landmarks are not annotated, and each face is annotated with at most 6 landmarks.
{\flushleft \textbf{LFPW:}} This dataset contains 1,132 training images and 300 testing images. The images in this dataset are given in the form of URLs, and some image links are no longer valid. We can only download 811 training images and 230 testing images.
{\flushleft \textbf{300W:}} The training set (3148 images) of this dataset is collected from the training sets of several exiting datasets, including LFPW (811), HELEN ~\cite{le2012interactive} (2000) and AFW (337). The full testing set is split into two subsets: (1) the common subset consists of the testing sets of LFPW (224) and HELEN (330), and (2) the challenging subset is composed of 135 images from IBUG~\cite{sagonas2013300}. All the images in this dataset are annotated with 68 facial landmarks.

\begin{figure*}[t]
\centering
    \includegraphics[width=2.05\columnwidth]{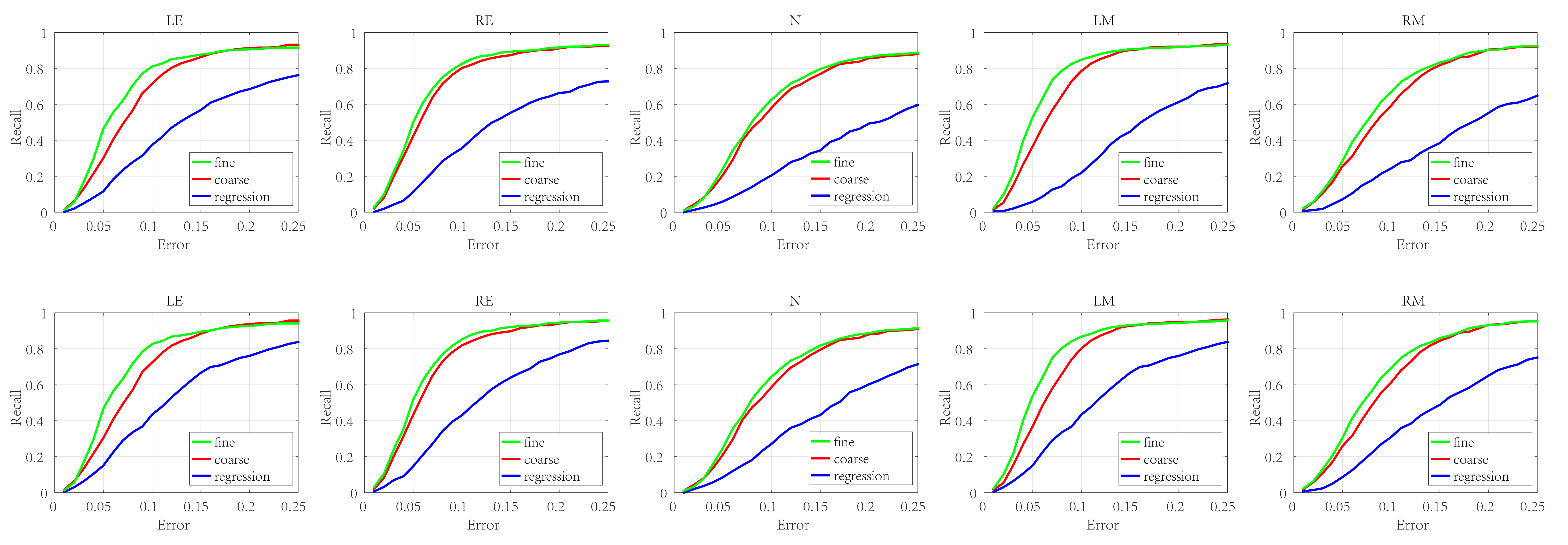}
\vspace{-8mm}
\caption{The recall of landmarks on AFW in unconstrained settings. The curves labeled ``fine'' and ``coarse'' show the performance of models with
and without branch networks, respectively. The curve labeled ``regression'' presents the performance of the regression network based on a single fully convolutional network. The top five figures demonstrate the recall performance when only 15 landmarks of each landmark type are predicted for each image, while the bottom five figures are the results with 30 predictive landmarks for each type of each image.}
\vspace{-1mm}
\label{fig:afw-unconstrained-evaluate}
\end{figure*}

\begin{figure*}[t]
  \centerline{
    \includegraphics[height=6.9cm]{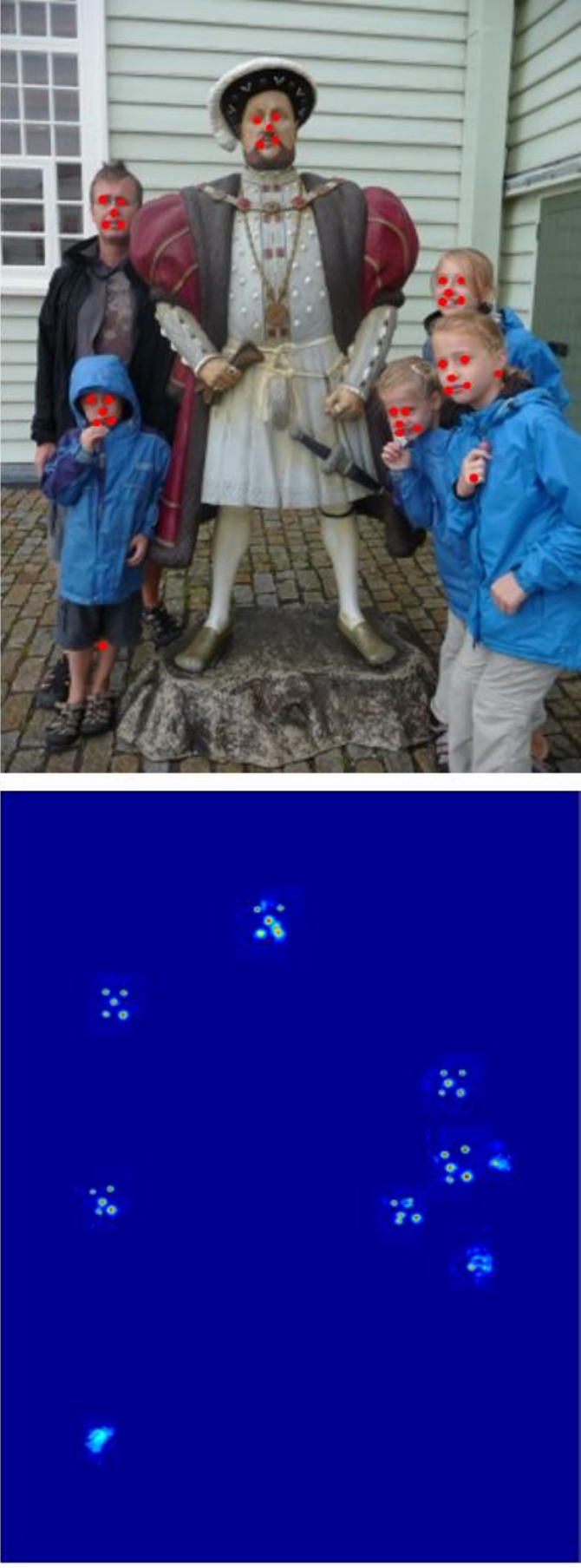}
    \includegraphics[height=6.9cm]{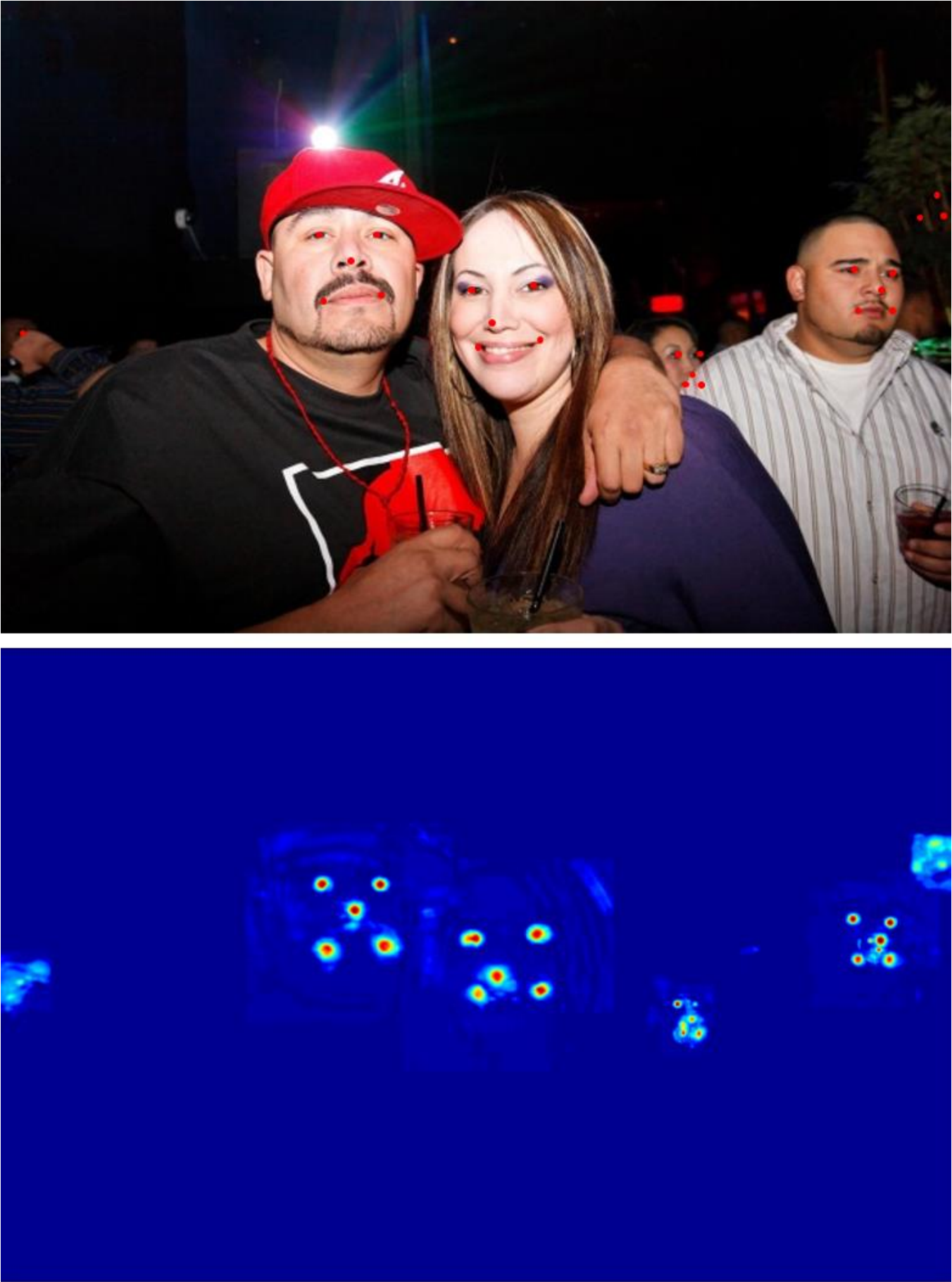}
    \includegraphics[height=6.9cm]{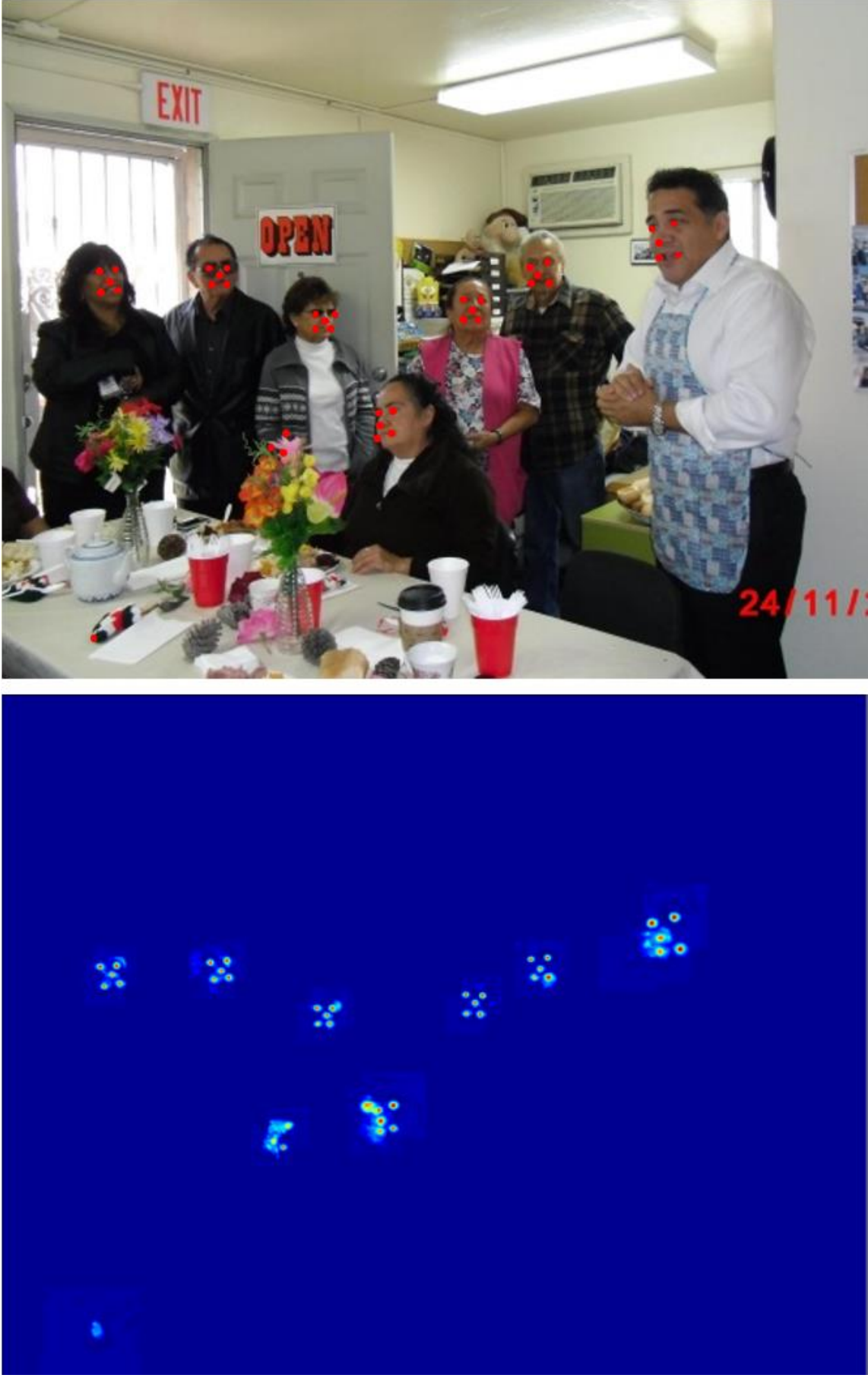}
    \includegraphics[height=6.9cm]{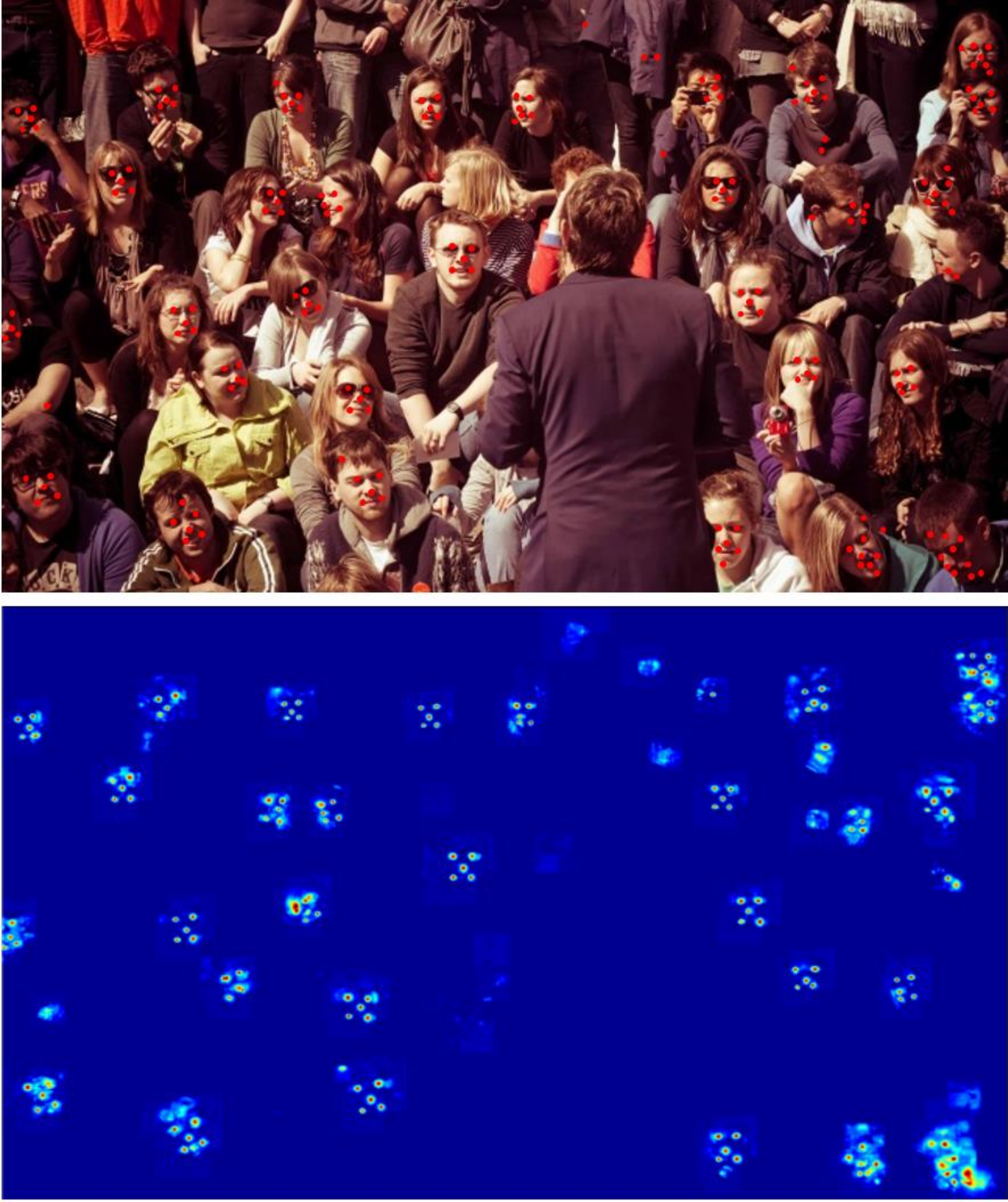}
    }
  \vspace{-2 mm}
  \caption{Qualitative facial landmark detection results in unconstrained settings. Our BB-FCN is capable of dealing with unconstrained facial images, even though the locations of facial regions and the number of faces in the image are unknown. Best viewed in color with zoom.}
  \vspace{-1mm}
\label{fig:test_landmark_in_the_wild}
\end{figure*}

\subsection{Evaluation Metric}
To evaluate the accuracy of facial landmark localization, we adopt the mean (position) error as the metric. For a specific type of landmark, the mean error is calculated as the mean distance between the detected landmarks of the given type in all testing images and their corresponding ground-truth positions, normalized with respect to the interocular distance. The (position) error of a single landmark is defined as follows:
\begin{equation}
\label{eq:mean_error}
err = \frac{\sqrt{(x - x')^2 + (y - y')^2}}{l} \times 100\%,
\end{equation}
where $(x, y)$ and $(x', y')$ are the ground-truth and detected landmark locations, respectively, and $l$ is the interocular distance. For the 300W dataset, the interocular distance is set to the Euclidean distance between the outer corners of two eyes, while for the other three landmark datasets, it is denoted as the Euclidean distance between the center points of the two eyes. In our experiments, we evaluate the mean error of every type of facial landmark and the average mean error over all landmark types, i.e., LE (left eye), RE (right eye), N (nose), LM (left mouth corner) and RM (right mouth corner), as well as A (average mean error of the five facial landmarks).

\subsection{Performance Evaluation for Unconstrained Settings}

\begin{table}[t]
\newcommand{\tabincell}[2]{\begin{tabular}{@{}#1@{}}#2\end{tabular}}
  \centering
  \caption{Average recalls of the complete backbone-branches network and the backbone network alone on AFW in unconstrained settings.
  PE refers to the acceptable position error.
  }
  \vspace{-3mm}
    \begin{tabular}{c|c|c|c|c}
    \hline
    \multirow{2}{*}{Model} &
    \multicolumn{2}{c|}{15 landmarks} &
    \multicolumn{2}{c}{30 landmarks} \\
    \cline{2-5}
    & PE=5\%  & PE=10\% & PE=5\% & PE=10\% \\
    \hline\hline
    backbone  & 31.1\% & 69.5\% & 31.5\% & 70.9\% \\
    \hline
    full model & 40.4\% & 75.6\% & 41.5\% & 77.6\% \\
    \hline
    \end{tabular}
  \vspace{0mm}
  \label{tab:coarse-vs-fine-unconstrained}
\end{table}

Our BB-FCN is capable of dealing with facial images taken in unconstrained settings, e.g., the locations of facial regions and the number of faces in the image are unknown.
In this setting, we use the recall-error curves to evaluate the performance of all comparative methods. A predictive facial landmark is considered to be correct if there exists a ground-truth landmark of the same type within the given position error. For a fixed number $m$ (such as 15 or 30) of predictive landmarks, the recall rate (the fraction of ground-truth annotations covered by predictive landmarks) varies as the acceptable position error increases; thus, a recall-error curve can be obtained.

To the best of our knowledge, very few facial landmark localization methods have been evaluated in the context of landmark detection under unconstrained settings. For fairness, we have also implemented a regression-based method using an FCN with nine convolutional layers, which can be expressed as follows:
$C(20, 5 \times 5 \times 3 )$ - $C(20, 5 \times 5 \times 20 )$ - $MP$ -
$C(30, 5 \times 5 \times 20 )$ - $C(30, 5 \times 5 \times 30 )$ - $MP$ -
$C(40, 5 \times 5 \times 30 )$ - $C(40, 5 \times 5 \times 40 )$ -
$C(30, 2 \times 2 \times 40 )$ - $C(30, 4 \times 4 \times 30 )$ - $C(15, 1 \times 1 \times 30)$.
With a training strategy similar to that of our backbone network, this regression-based network also takes a $32 \times 32$ image patch as input and generates a $15 \times 8 \times 8$ response map, each pixel of which corresponds to a $4 \times 4$ region of the input image. We formulate every three channels of the output response map as a group. Additionally, each pixel on a specific group indicates the probability of existence and the regressed two-dimensional location of the corresponding landmark type in a $4 \times 4$ region. During the testing phase, the same image pyramid is fed into the regression-based network for facial landmark inference.

We evaluate the performance of our BB-FCN and the regression-based deep model on the AFW dataset using an unconstrained setting. For those faces where one or both eyes are invisible, the interocular distances are set as 41.9\% of the length of their annotated bounding boxes\footnote{ The average ratio between the interocular distances of the common faces and the length of their annotated bounding boxes is 41.9\% on AFW}. Figure~\ref{fig:afw-unconstrained-evaluate} shows the recall-error curves of different types of landmarks, where the curves labeled ``fine'' and ``coarse'' illustrate the performance of our complete BB-FCN model and the single backbone network, respectively. The curve labeled ``regression'' indicates the performance of the above regression network based on a single FCN.

Our methods significantly outperform the regression network. With a prediction of 15 landmarks for each landmark type, the full model recalls 45\% more landmarks than the regression network when the acceptable position error is set within $8\%$ of the interocular distance.
Given more predicted landmarks, our complete BB-FCN model can achieve higher landmark recalls. As the number of landmark predictions of each type increases to $30$, the recalls of five landmarks within a position error of $25\%$ of the interocular distance are 94.1\%, 95.7\%, 91.5\%, 95.8\% and 95.2\%, respectively.
Meanwhile, the full model performs much better than the backbone network alone. The average recalls of five landmarks are shown in Table~\ref{tab:coarse-vs-fine-unconstrained}, which shows that the full model improves the recall rate by approximately 10\% and 6\% when the acceptable position error is set as 5\% or 10\%, respectively.
As shown in Figure~\ref{fig:test_landmark_in_the_wild}, our BB-FCN can generate high-quality heat maps and detect almost all the facial landmarks, even though some false positives exist. These false positives are some tiny and blurry regions (such as treetops and hands) that have rich texture or have similar shapes and colors as faces.

\subsection{Performance Evaluation for Constrained Settings}
In this setting, because the face bounding boxes are given, we can directly feed the face regions into our BB-FCN network to locate the facial landmarks. We will compare our method with state-of-the-art methods on the five landmark types and on dense landmark types.

\begin{figure}[t]
\begin{center}
   \includegraphics[width=0.95\columnwidth,height=6.5cm]{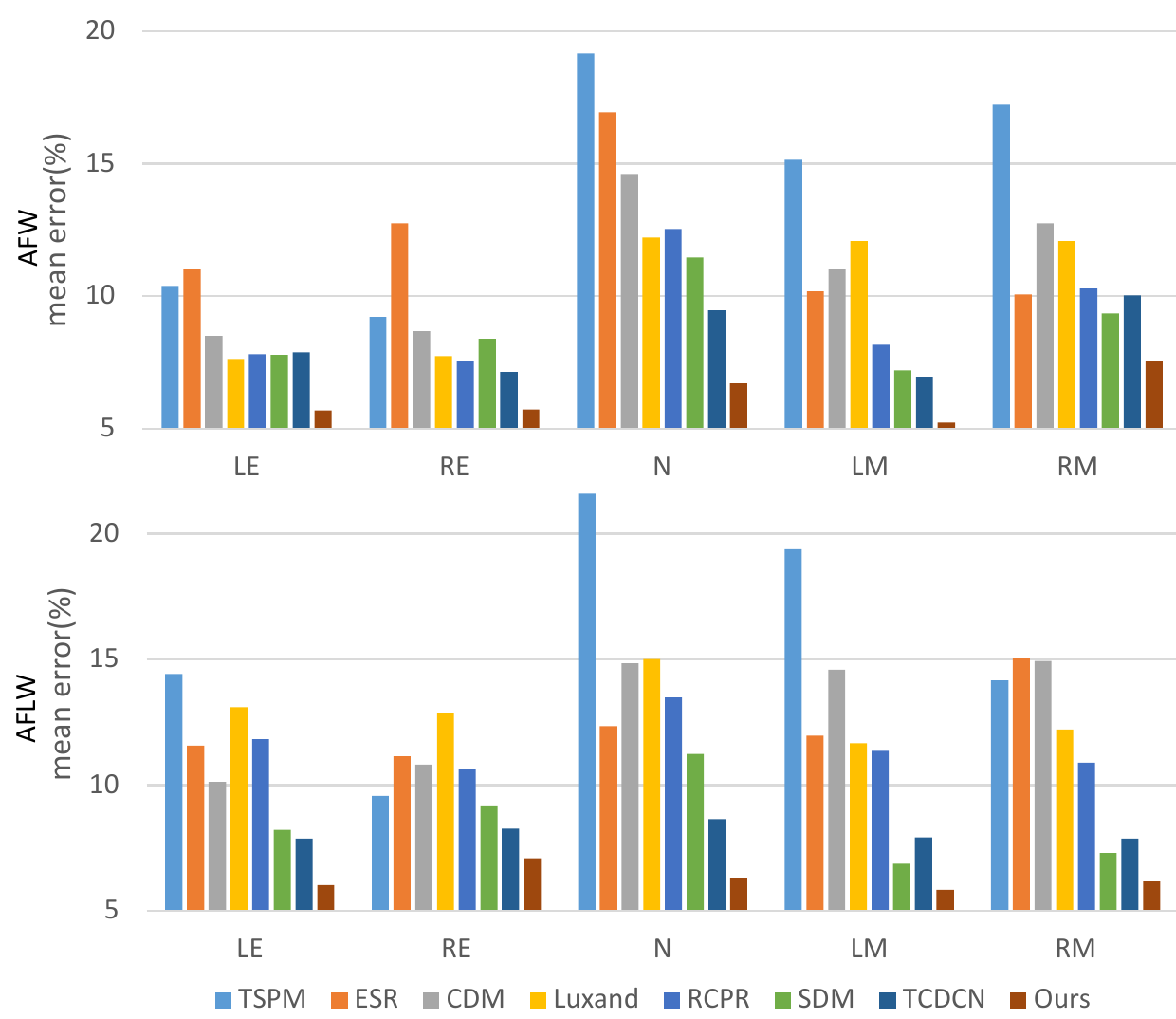}
\end{center}
\vspace{-4mm}
   \caption{Comparisons with state-of-the-art methods on two public datasets. The top row shows the corresponding results on AFW, and the bottom row shows the corresponding results on AFLW. The average mean errors of all considered methods are summarized in Table \ref{tab:ave-mean-error}.}
\vspace{-3mm}
\label{fig:mean-error}
\end{figure}

\begin{table}[t]
\newcommand{\tabincell}[2]{\begin{tabular}{@{}#1@{}}#2\end{tabular}}
  \centering
  \caption{Average mean errors of our method and of all other competing methods on AFW and AFLW.}
  \vspace{-1mm}
    \begin{tabular}{c|c|c}
    \hline
    \tabincell{c}{Dataset } & \tabincell{c}{ AFW} & \tabincell{c}{AFLW} \\
    \hline\hline
    TSPM & 14.31 & 15.9 \\
    \hline
    ESR & 12.2 & 13 \\
    \hline
    CMD  & 11.1 & 13.1 \\
    \hline
    Luxand & 10.4 & 12.4 \\
      \hline
    RCRR & 9.3 & 11.6 \\
      \hline
    SDM  & 8.8 & 8.5 \\
      \hline
    TCDCN  & 8.2 & 8.0 \\
      \hline
    RAR & - & 7.23 \\
      \hline
    MTCNN  & - & 6.9 \\
      \hline
    UD & - & 6.58 \\
      \hline
    Ours  & 6.18 & 6.28 \\
      \hline
    \end{tabular}
  \vspace{0mm}
  \label{tab:ave-mean-error}
\end{table}

\subsubsection{\bf{Evaluation on Five Landmark Types}}
We compare our method with other state-of-the-art methods, i.e., \footnote{Some results on AFW and AFLW are quoted from~\cite{zhang2014facial}.} robust cascaded pose regression (RCPR)~\cite{burgos2013robust}, tree structured part model (TSPM) ~\cite{zhu2012face}, Luxand face SDK~\footnote{Luxand face SDK: http://www.luxand.com/}, explicit shape regression (ESR)~\cite{cao2014face}, cascaded deformable shape model (CDM)~\cite{yu2013pose}, supervised descent method (SDM)~\cite{xiong2013supervised}, tasks-constrained deep convolutional network (TCDCN)~\cite{zhang2014facial}, multitask cascaded convolutional networks (MTCNN)~\cite{zhang2016joint}, recurrent attentive-refinement networks (RAR)~\cite{xiao2016robust}, and unsupervised discovery (UD)~\cite{zhang2018unsupervised}.

\begin{figure*}
\begin{center}
 \includegraphics[width=1.85\columnwidth]{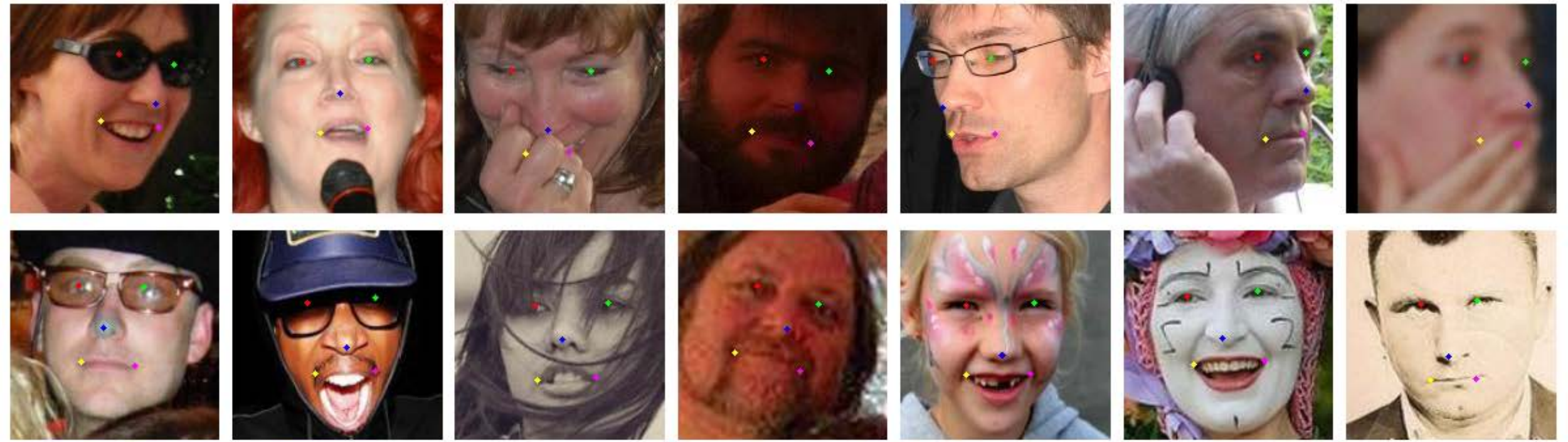}\vspace{-4mm}
\end{center}
  \vspace{-2mm}
   \caption{Qualitative facial landmark localization results of our method. The first row shows the result on AFW, while the second row shows the result on AFLW. Our method is robust under occlusions, exaggerated expressions and extreme illumination.}
 \vspace{0mm}
\label{fig:landamrk_visual}
\end{figure*}

On the AFW dataset, our average mean error over the five landmark types is 6.18$\%$, which is an improvement over the performance of the state-of-the-art TDCN by 24.6$\%$. On the AFLW dataset, our BB-FCN model achieves 6.28$\%$ average mean error, a 21.5$\%$ improvement over TDCN. Figure~\ref{fig:mean-error} and Table~\ref{tab:ave-mean-error} demonstrate that our BB-FCN network outperforms all competing methods on the three datasets. The qualitative results presented in Figure~\ref{fig:landamrk_visual} show that our method is robust under occlusions, exaggerated expressions and extreme illumination.

\begin{table}[t]
\newcommand{\tabincell}[2]{\begin{tabular}{@{}#1@{}}#2\end{tabular}}
  \centering
\caption{Average mean errors of landmark detection on the 300W dataset.}
  \vspace{-3mm}
    \begin{tabular}{c|c|c|c}
    \hline
    \tabincell{c}{ Methods } & \tabincell{c}{ Common Set} & \tabincell{c}{ Challenging Set } & \tabincell{c}{ Full Set }  \\
    \hline\hline
    TSPM & 8.22 & 18.33 & 10.22 \\
    \hline
    RCPR & 6.18 & 17.26 & 8.35 \\
    \hline
    SDM  & 5.57 & 15.40 & 7.50 \\
    \hline
    ESR  & 5.28 & 17.00 & 7.58 \\
    \hline
    LBF  & 4.95 & 11.98 & 6.32 \\
    \hline
    CFSS & 4.73 & 9.98 & 5.76 \\
    \hline
    CFAN  & 5.50 & 16.78 & 7.69 \\
    \hline
    3DDFA & 6.15 & 10.59 & 7.01 \\
    \hline
    3DDFA+SDM & 5.53 & 9.56 & 6.31 \\
    \hline
    TCDCN & 4.8 & 8.6 & 5.54 \\
    \hline
    RAR & 4.12 & 8.35 & 4.94 \\
    \hline
    Pose-Invariant   & 5.43 & 9.88 & 6.30\\
    \hline
    RDR              & 5.03 & 8.95 & 5.80\\
    \hline
    Two-Stage$_{OD}$ & 4.36 & 7.56 & 4.99\\
    \hline
    RCN${^+}$        & 4.20 & 7.78 & 4.90\\
    \hline
    Ours & \textbf{3.85} & \textbf{7.50} & \textbf{4.56} \\
    \hline
    \end{tabular}
  \label{tab:300W_result}
\end{table}

\begin{figure*}
\begin{center}
 \includegraphics[width=1.85\columnwidth]{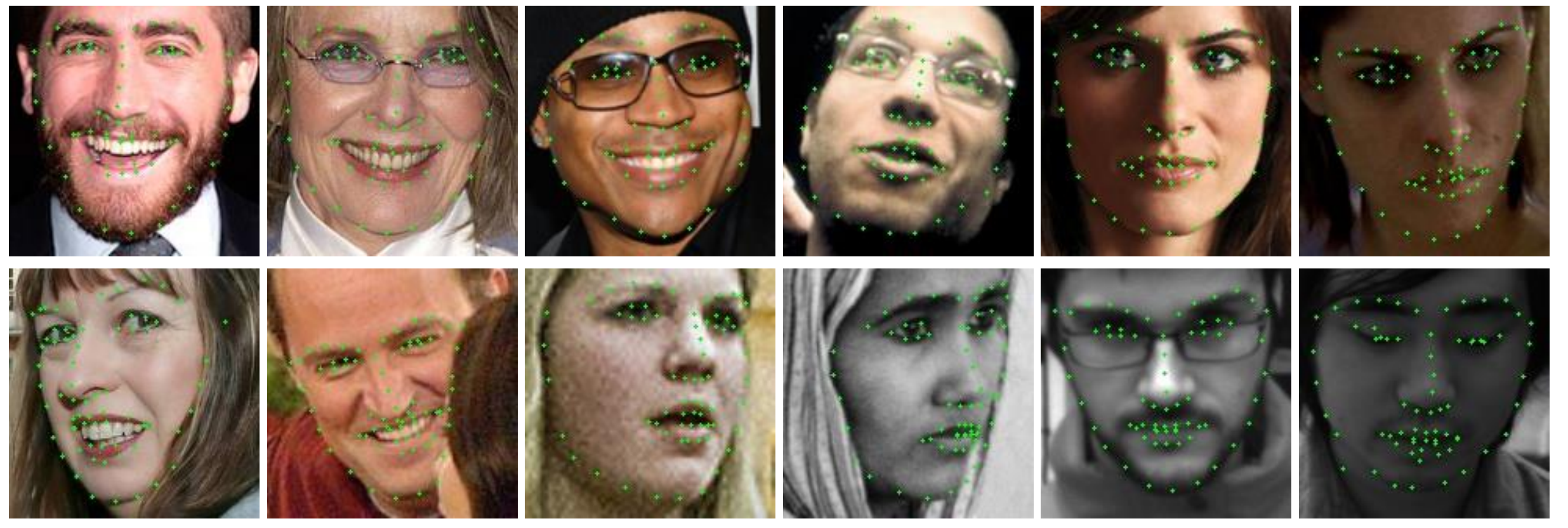}
 \vspace{-6mm}
\end{center}
   \caption{Qualitative facial landmark localization results of our method on the 300W dataset. The first row shows the result on the common set, while the second row demonstrates the result on the challenging set.}
\vspace{-0mm}
\label{fig:300w_visual}
\end{figure*}

\subsubsection{\bf{Evaluation on Dense Landmark Types}}

We can use our BB-FCN network for dense landmark prediction by simply extending the number of branches in the branch network. We evaluate our extended method on LFPW with 29 landmarks and on 300W with 68 landmarks. Because dense landmark prediction requires more facial details to distinguish landmarks with similar appearances, such as left-eyebrow-center-top and left-eyebrow-center-bottom, we enlarge the input images of BB-FCN to $64 \times 64$. Due to the differences between the landmark types of our collected dataset and LFPW, we fine-tune the network using the training set of LFPW. Moreover, for the 68 landmark types, we train our network from scratch with the training set of the 300W dataset.

We compare our method with other state-of-the-art methods on the LFPW dataset. The other methods include consensus of exemplars (CE)~\cite{belhumeur2013localizing}, explicit shape regression (ESR)~\cite{cao2014face} and ensemble of regression trees (ERT)~\cite{kazemi2014one}. Table~\ref{tab:ave-mean-error-LFPW} shows that our BB-FCN achieves 3.35$\%$ average mean error, outperforming the other three state-of-the-art methods.

We also compare the performance of our proposed method with the results of other state-of-the-art methods on the 300W testing set with 68 landmarks. The first class of compared methods are cascaded regression-based models, including TSPM, RCPR, SDM, ESR, LBF~\cite{ren2014face} and CFSS~\cite{zhu2015face}. The second class are deep-learning-based methods, including TCDCN, 3DDFA~\cite{zhu2016face}, CFAN~\cite{zhang2014coarse}, RAR~\cite{xiao2016robust}, Pose-Invariant~\cite{jourabloo2017pose}, RDR~\cite{xiao2017recurrent}, Two-Stage$_{OD}$~\cite{lv2017deep}, and RCN${^+}$~\cite{honari2018improving}. As shown in Table~\ref{tab:300W_result},  our proposed method significantly outperforms all the other state-of-the-art methods across all different testing sets; specifically, our complete model lowers the average mean error achieved by the best-performing existing algorithm~(RCN${^+}$) by 8.3\%, 3.6\% and 6.9\% on the common set, the challenging set and the full set, respectively. 
Figure~\ref{fig:300w_visual} presents some example results of our proposed pixel-labeling method for dense landmark prediction.

\begin{table}[t]
\newcommand{\tabincell}[2]{\begin{tabular}{@{}#1@{}}#2\end{tabular}}
  \centering
  \caption{Average mean errors of our method and all other competing methods on LFPW. }
  \vspace{-2mm}
    \begin{tabular}{c|c|c|c|c}
    \hline
    \tabincell{c}{Methods } & \tabincell{c}{CE} & \tabincell{c}{ ESR} & \tabincell{c}{ERT} & \tabincell{c}{Ours} \\
    \hline\hline
    Error & 4.00 & 3.43 & 3.80 & 3.35 \\
    \hline
    \end{tabular}
  \vspace{-2mm}
  \label{tab:ave-mean-error-LFPW}
\end{table}

\subsection{Ablation Study}
\begin{figure}[t]
\centering
   \includegraphics[width=0.9\columnwidth]{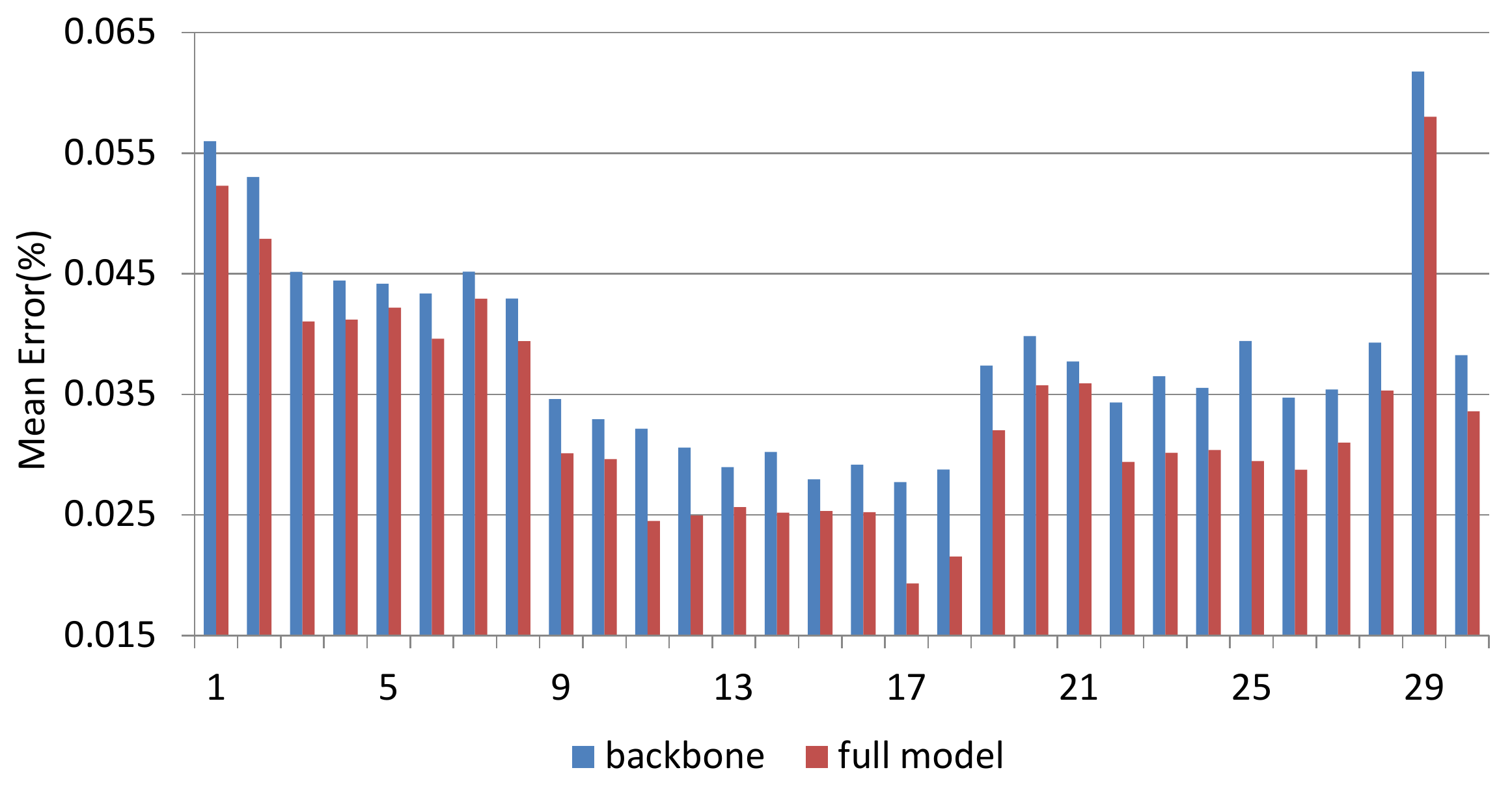}
\vspace{-4mm}
   \caption{Performance evaluation of the complete backbone-branches network and the backbone network alone on LFPW. The mean error of every type of landmark is decreased to a certain degree when the branch networks are used. The 30th column is the average mean error. }
\vspace{-0mm}
\label{fig:LFPW_self_comparison}
\end{figure}

\begin{figure}[t]
\centering
   \includegraphics[width=0.9\columnwidth]{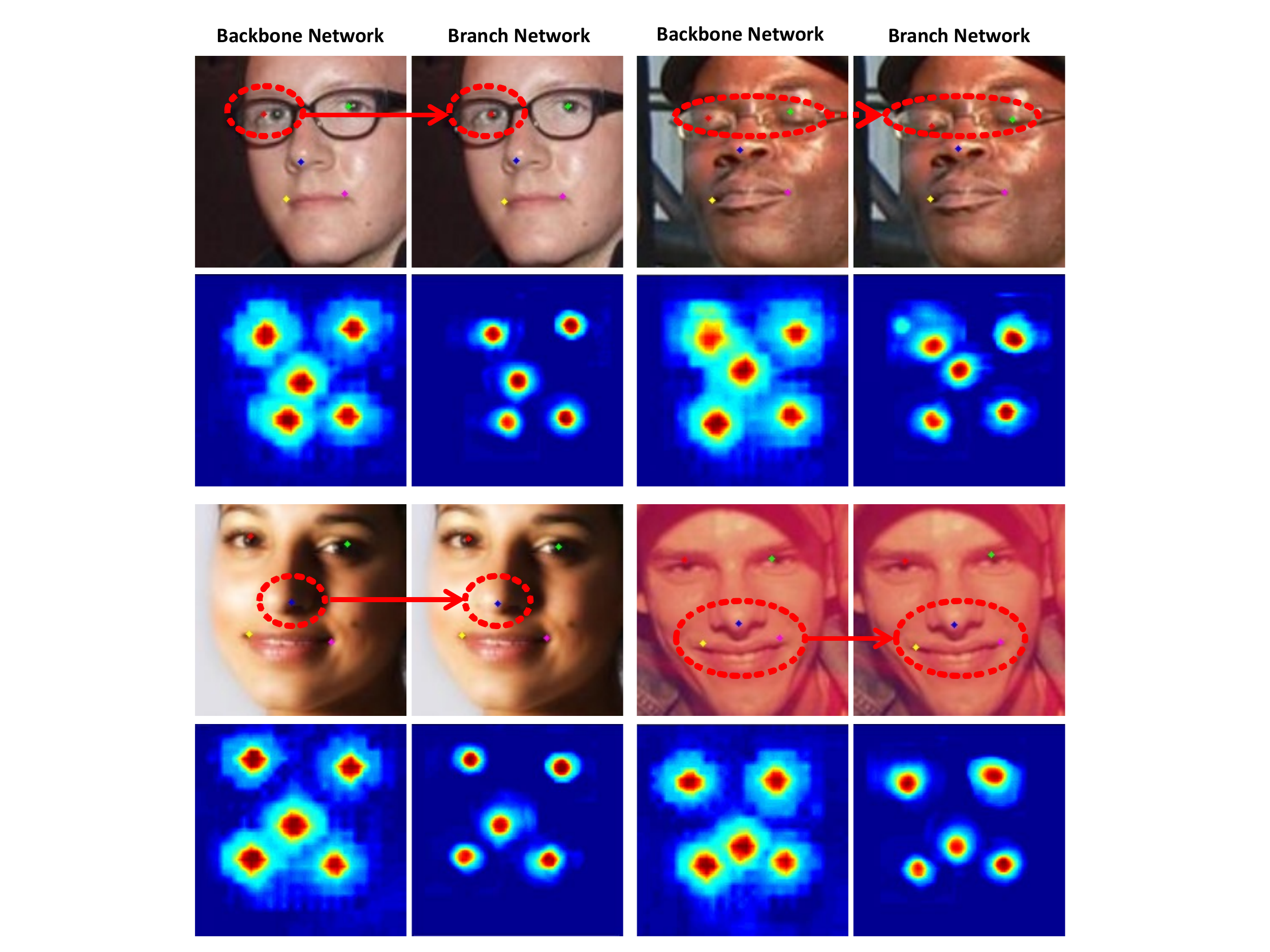}
\vspace{-3mm}
   \caption{Examples of improvements made by the branch networks. The response heat maps of the branch networks are more compact and precise. Best viewed in color.}
\vspace{-3mm}
\label{fig:BB-FCN-improvement}
\end{figure}

Our proposed BB-FCN is composed of two components: the backbone network and the branch networks. To show the effectiveness and necessity of these two components, we compare the landmark prediction results produced by the single backbone network with those of the complete BB-FCN network. As shown in Table~\ref{tab:coarse-vs-fine}, the average mean error on AFLW is decreased from 8.31$\%$ to 6.28$\%$, with an approximately 24.4$\%$ relative improvement, after the branch networks are added to perform landmark refinement. The quantitative comparison shown in Figure~\ref{fig:LFPW_self_comparison} further demonstrates that the prediction error of every type of facial landmark enjoys a varying degree of reduction on LFPW. Figure~\ref{fig:BB-FCN-improvement} shows the visual improvements achieved with the branch networks over the single backbone network. As shown, the output heat maps of the branch networks are more compact and precise than those of the backbone network, which can well explain the better performance of branch networks.

\begin{table}[t]
\newcommand{\tabincell}[2]{\begin{tabular}{@{}#1@{}}#2\end{tabular}}
  \centering
  \caption{Average mean errors of the complete backbone-branches network and the backbone network on AFW and AFLW.
  }
  \vspace{-3mm}
    \begin{tabular}{c|c|c|c|c}
    \hline
    \multirow{2}{*}{landmark  type} &
    \multicolumn{2}{c|}{AFW} &
    \multicolumn{2}{c}{AFLW} \\
    \cline{2-5}
    & backbone  & full model & backbone & full model \\
    \hline\hline
    LE & 7.02 & 5.69 & 9.46 & 6.02 \\
    \hline
    RE & 6.79 & 5.72 & 8,60 & 7.08 \\
    \hline
    N  & 8.35 & 6.71 & 8.39 & 6.31 \\
      \hline
    LM & 7.11 & 5.22 & 7.40 & 5.83 \\
      \hline
    RM & 7.98 & 7.58 & 7.73 & 6.15 \\
      \hline
    A  & 7.45 & 6.18 & 8.31 & 6.28 \\
    \hline
    \end{tabular}
  \vspace{-5mm}
  \label{tab:coarse-vs-fine}
\end{table}

\subsection{The Effectiveness of Face Proposal Generation}
In this experiment, we demonstrate the effectiveness of our landmark prediction network in face proposal generation. A predicted facial landmark typically indicates the existence of a face; therefore, we can generate face proposals from the response heat map of the BB-FCN. For a type-$k$ predicted facial landmark at level $l$, we generate a $32 \times 32$ face candidate window centered at the landmark location from the RGB image at the $l^{th}$ level of the pyramid.
 We then apply NMS to face proposals generated using each type of landmark. After fine-tuning the location and edge length of face proposals with Net-12 (the first network of cascade CNN~\cite{li2015convolutional}), we apply NMS to all face proposals again.

We compare our method with three generic object proposal generators~\cite{zitnick2014edge,uijlings2013selective,arbelaez2014multiscale} and a face-specific proposal generator, Faceness~\cite{Yang2015From}, on FDDB.
For a fair comparison, following~\cite{Yang2015From}, we also transform the original ground-truth ellipses in FDDB into minimal rectangular bounding boxes.
Table~\ref{tab:proposal_result}\footnote{The results from the compared methods are quoted from~\cite{Yang2015From}.} shows that our method achieves high recalls using a very small number of face proposals due to the accuracy of landmark localization. Our method can detect 72.8$\%$ of faces using only two proposals per image and 81.2$\%$ of faces using three proposals per image on FDDB. It detects 91.5$\%$ of faces when at most 20 face proposals are generated from each image.
With a similar proposal generation strategy as our method, Faceness~\cite{Yang2015From} utilizes facial attributes to calculate the facial part response maps and then generates the region proposal from these response maps.
Compared with~\cite{Yang2015From}, our method can generate more accurate landmark (part) response maps by explicitly locating the facial landmarks, and it is more robust to overcome partial occlusions and head pose variations.

\begin{table}[t]
\newcommand{\tabincell}[2]{\begin{tabular}{@{}#1@{}}#2\end{tabular}}
  \centering
\caption{The number of proposals needed for different recall rates on FDDB.}
  \vspace{-3mm}
    \begin{tabular}{c|c|c|c|c}
    \hline
    \tabincell{c}{Proposal method } & \tabincell{c}{75\%} & \tabincell{c}{80\%} & \tabincell{c}{85\%} & \tabincell{c}{90\%} \\
    \hline\hline
    EdgeBox & 132 & 214 & 326 & 600 \\
    \hline

    MCG & 191 & 292 & 453 & 941 \\
    \hline
    Selective Search & 153 & 228 & 366 & 641 \\
    \hline
    EdgeBox+Faceness & 21 & 47 & 99 & 288 \\
    \hline
    MCG+Faceness & 13 & 23 & 55 & 158 \\
    \hline
    Selective Search+Faceness & 24 & 41 & 91 & 237 \\
    \hline
    Ours & $ > $2 & $ < $3 & 5 & 9 \\
    \hline
    \end{tabular}
  \vspace{-3mm}
  \label{tab:proposal_result}
\end{table}

\subsection{Evaluation on Face Detection Performance}
Our BB-FCN network can locate various landmarks in unconstrained settings and generate high-quality face proposals, which can enhance the performance of existing face detectors, such as cascade CNN~\cite{li2015convolutional}, particularly under severe occlusions and large pose variation. 
Cascade CNN is one of the up-to-date fast face detectors. It relies on six cascaded convolutional neural networks to locate faces in an image. We retrain this detector using our collected landmark dataset and Pascal VOC 2012, and we achieve similar performance on FDDB. We replace the original face proposals used by cascade CNN with our landmark-based proposals. All other parts of the method remain the same. The experimental results indicate that the modified cascade CNN achieves state-of-the-art performance on two public face detection benchmarks: FDDB and AFW.

\begin{figure*}
\centering
{\includegraphics[width=0.49\linewidth]{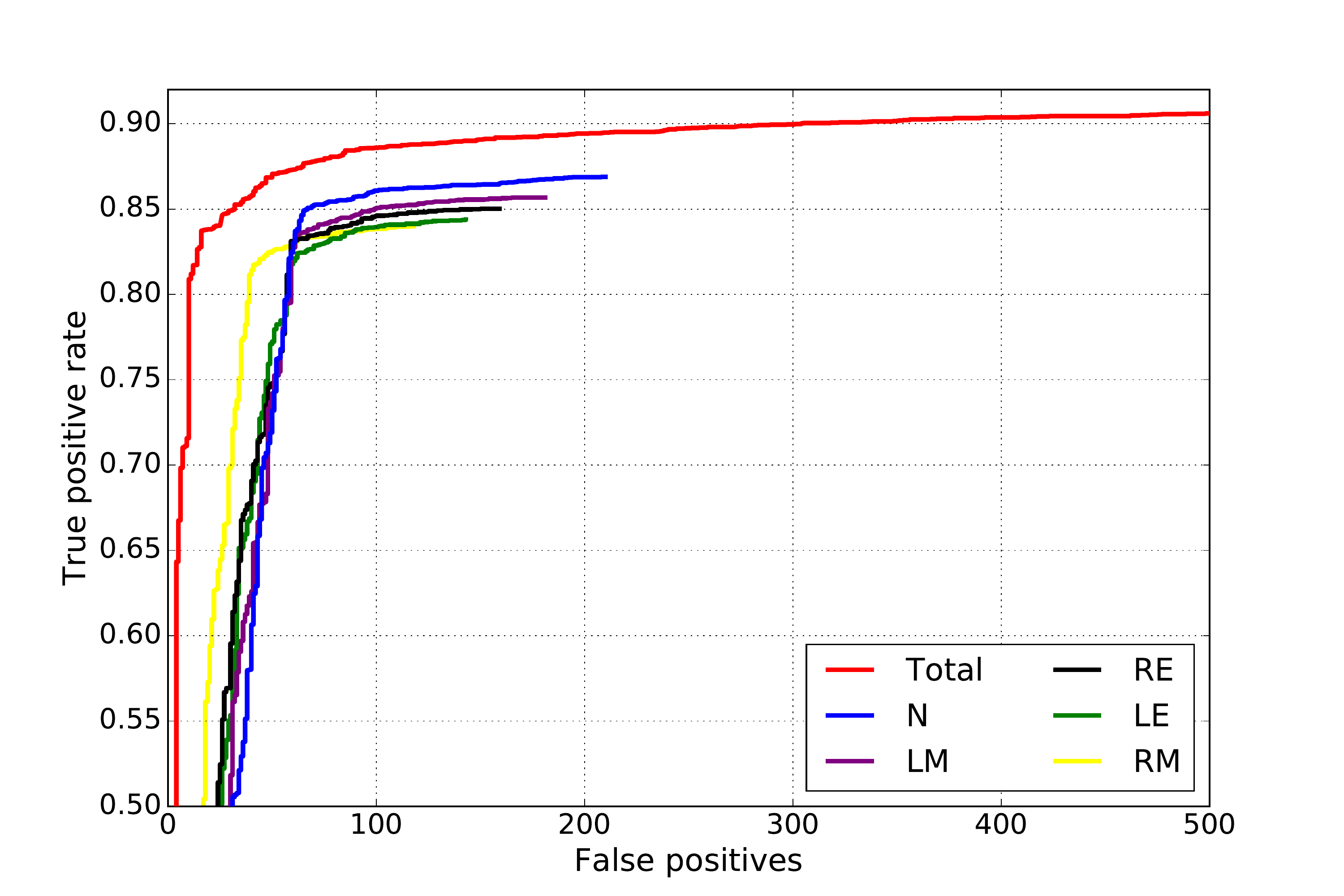}}
{\includegraphics[width=0.49\linewidth]{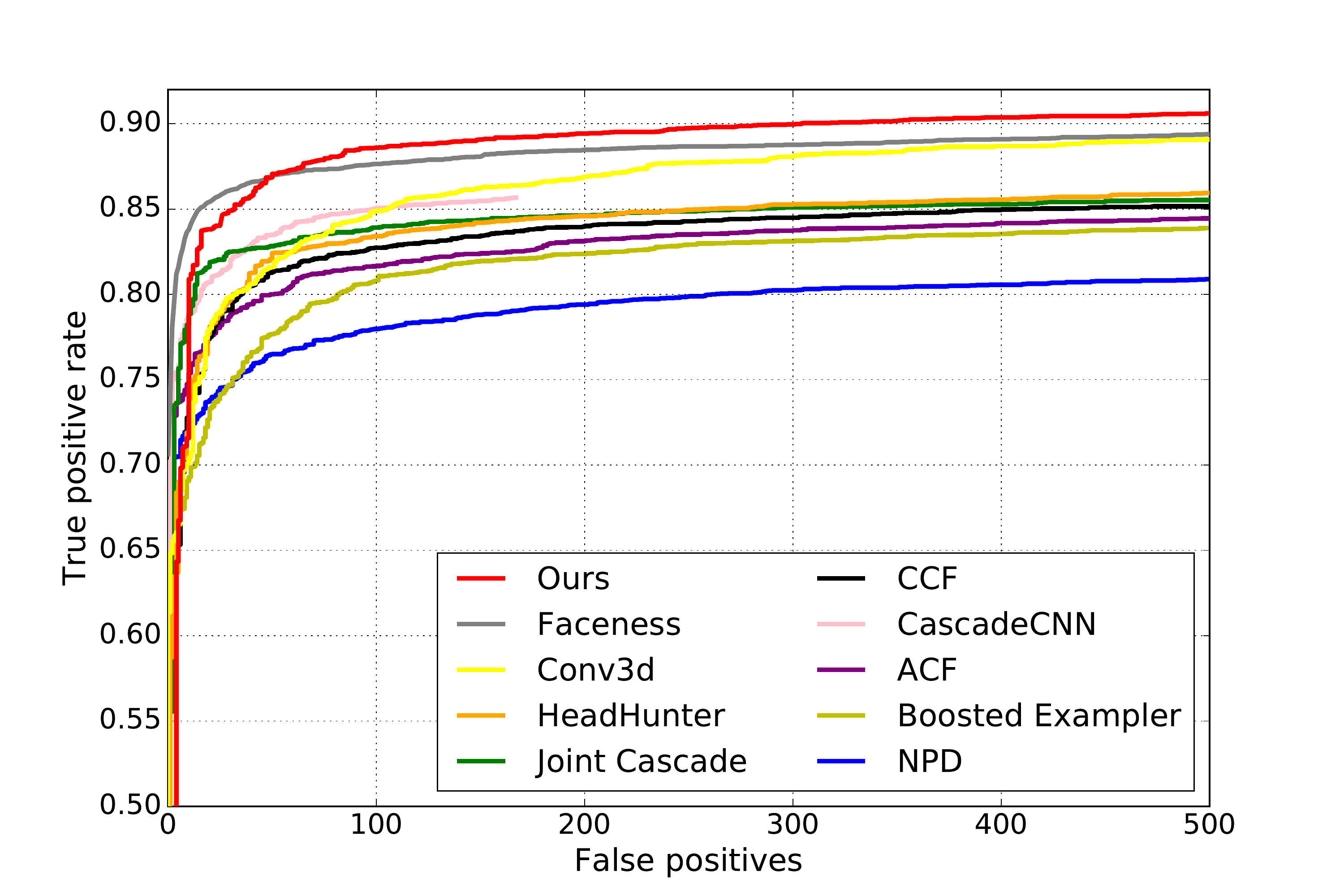}}
\vspace{-2mm}
\caption{Left: Face proposals induced by different landmark types exhibit different levels of effectiveness in face detection. Using face proposals induced by all five landmark types significantly improves the performance achieved with individual landmark types. Right: On the FDDB dataset, we compare our method against other state-of-the-art methods. When the number of false positives is fixed at 350, the recall achieved with our method is 90.17\%, which is higher than all other methods. }
\vspace{-2mm}
\label{fig:roc-fddb}
\end{figure*}

\subsubsection{FDDB}
As a large-scale face detection benchmark, FDDB contains 5,171 annotated faces in 2,845 images. It uses elliptic face annotations and defines two types of evaluations: the discontinuous score and continuous score. We use the discontinuous score evaluation, which counts the number of detected faces versus the number of false alarms. A detected bounding box is taken as the true positive only if the IoU between this bounding box and the bounding box of a ground-truth face is above 0.5. We uniformly enlarge our square bounding boxes vertically by 25\% to better approximate elliptic annotations in FDDB.

As shown in Figure~\ref{fig:roc-fddb}, face proposals defined by different landmark types exhibit different levels of effectiveness in face detection. The nose landmark achieves the best performance among all landmark types. Using face proposals defined by all five landmark types significantly improves the performance achieved with individual landmark types. 

We compare our method with nine recently published state-of-the-art methods on the FDDB dataset. These methods include cascade CNN~\cite{li2015convolutional}, Faceness~\cite{Yang2015From}, CCF~\cite{yang2015convolutional}, Conv3d~\cite{li2016face}, HeadHunter~\cite{mathias2014face}, joint cascade~\cite{chen2014joint}, boosted exemplar~\cite{li2014efficient}, ACF~\cite{yang2014aggregate} and NDP~\cite{li2014efficient}. Figure~\ref{fig:roc-fddb} shows that our method outperforms all nine state-of-the-art methods by a considerable margin. When the number of false positives is fixed at 167, our method achieves a significant margin of 3.51\% in recall rate over the baseline cascade-CNN~\cite{li2015convolutional}. When the number of false positives is fixed at 350, our method achieves a 90.17\% recall rate, which is higher than the 88.92\% recall rate achieved by Faceness~\cite{Yang2015From}.
When the number of false positives increases to 500, our method obtains a recall rate of 90.6\% with at most 20 face proposals per images. In contrast, when trained with approximately 83K face images, joint training cascade CNN~\cite{qin2016joint} generates nearly 1000 proposals on average before applying the MNS and only obtains a recall of 88.2\% with 1000 false positives. Recently, SAFD~\cite{hao2017scale} trained their network with 350K private face images and obtained a recall of 93.8\% with 1000 false positives. However, our method can achieve competitive performance with only 16K face images in our SYSU16K dataset.

\subsubsection{AFW}
\begin{figure}[t]
\centering
   \includegraphics[width=0.8\columnwidth]{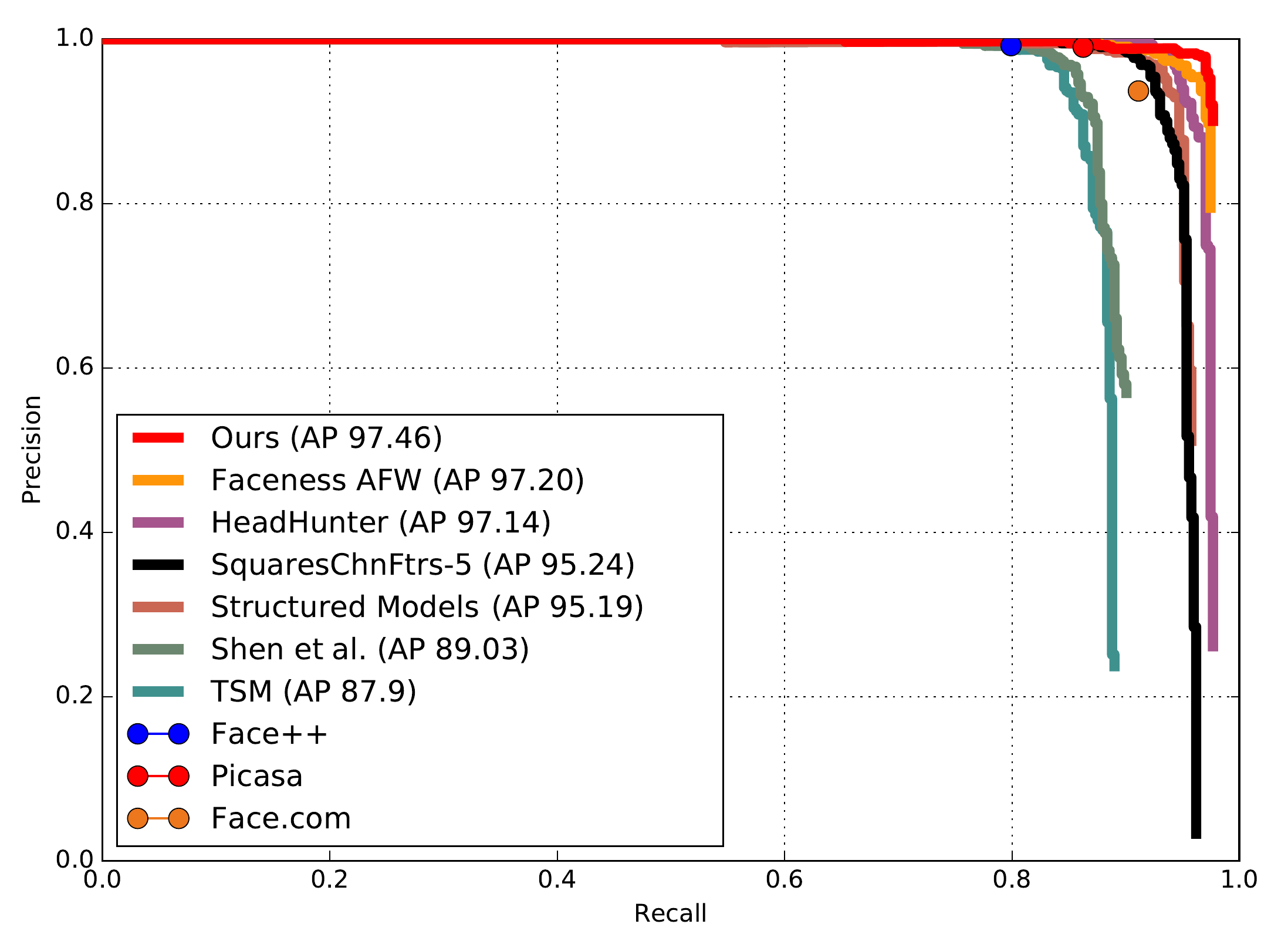}
\vspace{-5mm}
   \caption{Precision-recall curves of 10 face detection methods on the AFW dataset. The performance of our face detector is comparable to that of other state-of-the-art techniques.}
\label{fig:AFW_PR}
\vspace{-3mm}
\end{figure}

We adopt the precision-recall protocol when performing evaluation on the AFW dataset. We compare our method with Faceness~\cite{Yang2015From}, HeadHunter~\cite{mathias2014face}, structured models~\cite{yan2014face}, SquareChnFtrs-5\cite{mathias2014face}, Shen et al.~\cite{shen2013detecting}, TSM~\cite{zhu2012face}, Face.com, Face++ and Picasa. As shown in Figure~\ref{fig:AFW_PR}, with an average precision of 97.46\%, the performance of our detector is comparable to that of other state-of-the-art techniques. 

\subsection{Limitations}
\begin{figure}[!htbp]
\centering
   \includegraphics[width=0.85\columnwidth]{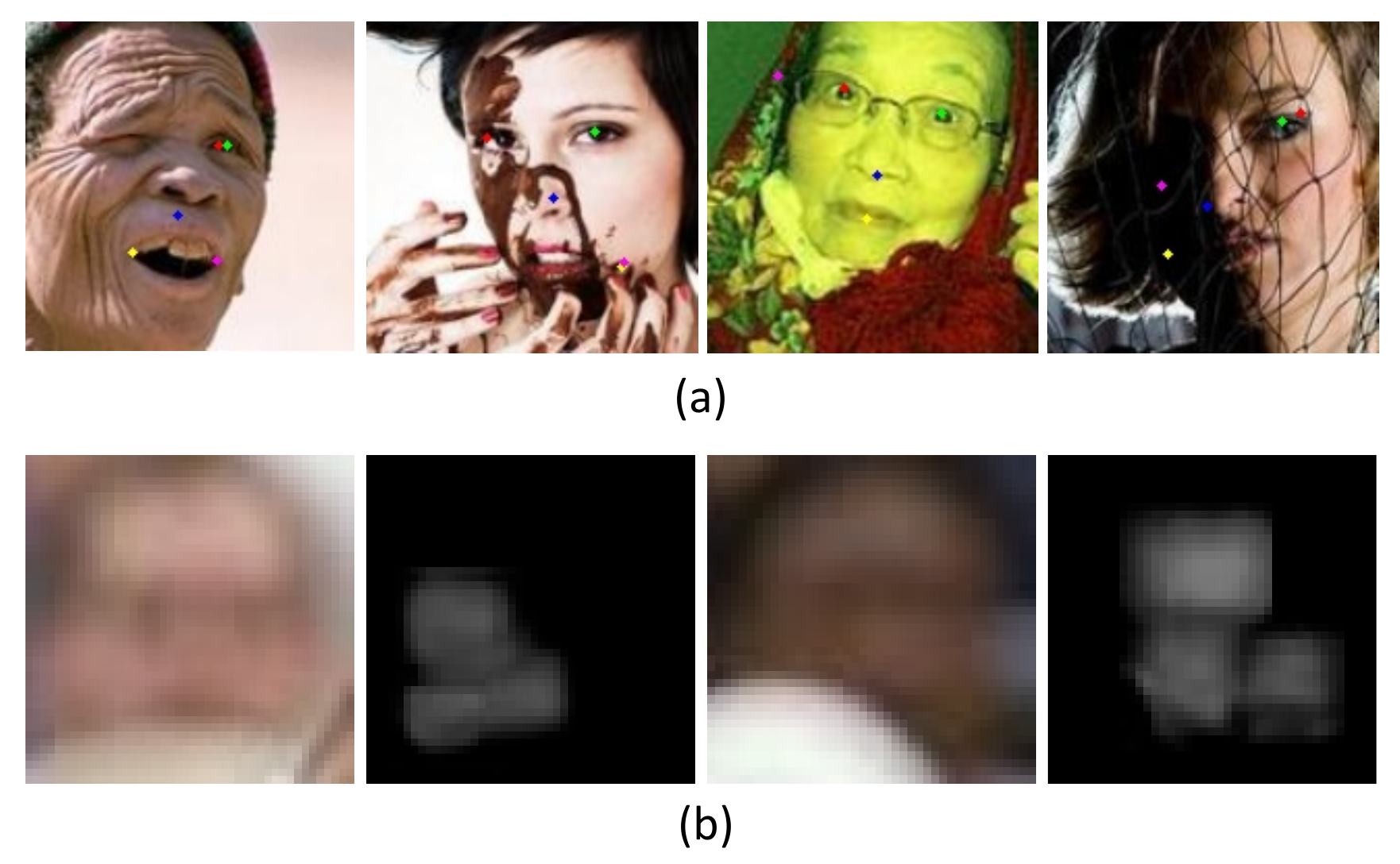}
\vspace{-5mm}
   \caption{Failure cases of our BB-FCN network. (a) Incorrect landmark prediction results that violate the normal spatial layout of human facial landmarks. (b) Two blurry faces from FDDB and their response heat maps.}
\vspace{-4mm}
\label{fig:limitation}
\end{figure}

In this section, we present failure cases of our BB-FCN network. In our experiments, we found that BB-FCN occasionally generates results that do not conform to the normal spatial layout of human facial landmarks, as shown in Figure~\ref{fig:limitation}(a). The main reason for this phenomenon is the lack of constraints on relative landmark positions in the loss function. Second, BB-FCN fails to highlight facial landmarks in blurry images, as shown in Figure~\ref{fig:limitation}(b). This negatively impacts the performance of our face proposal method on FDDB, which contains many blurry faces.

\subsection{Runtime Efficiency}
One of the most important characteristics of our landmark and face detectors is the efficiency. Our method achieves practical runtime efficiency via a coarse-to-fine pipeline. Table~\ref{tab:landmark_detection_time} shows the running times of
several deep models for five facial landmark detection under constrained settings. Among these models, TCDCN requires 18~ms to process a facial image on an Intel Core i5 CPU, which is 7 times faster than CDCN~\cite{sun2013deep}. CFAN~\cite{zhang2014coarse} costs 30~ms to run multiple autoencoders. Our method only needs 9~ms on an Intel Core i5 2.80~GHz CPU and 1.8~ms on an NVIDIA Titan X GPU.  For the localization of 68 landmarks, our method costs 10~ms to process a face region on the same GPU.

For the unconstrained setting, to locate the landmarks of the tiny faces for a high recall rate, we build a 20-level image pyramid on the AFW and FDDB datasets, and our landmark network runs at approximately 6 PFS on the same GPU. However, the level number of the image pyramid can be dynamically adjusted based on the acceptable minimum face size. For example, to locate the landmarks of faces with sizes larger than $80\times 80$ from $640\times 480$ VGA images, we only need to build an image pyramid with 7 levels. In this case, our landmark networks can run at 30 FPS, while our face detection pipeline can run at approximately 20 FPS on the same GPU thanks to our efficient proposal generator and the cascade CNN detector.
For comparison, Shen et al.~\cite{shen2013detecting} process a 1280-pixel wide image in less than 10 seconds and DP2MFD~\cite{ranjan2015deep} runs at 0.285 FPS on an Nvidia Tesla K20, while the ResNet101-based detector proposed by HR~\cite{hu2017finding} runs at 3.1 FPS on 720p resolution. With a similar speed as our network, Faceness~\cite{Yang2015From} can process a VGA image within 50 ms on a Titan Black GPU, but their performance is worse than ours.

\begin{table}[t]
\newcommand{\tabincell}[2]{\begin{tabular}{@{}#1@{}}#2\end{tabular}}
  \centering
  \caption{Comparison of Running Times on CPU among Deep Models for Five Facial Landmark Detection. }
  \vspace{1mm}
  \resizebox{0.25\textwidth}{!}
{
    \begin{tabular}{c|c}
    \hline
    \tabincell{c}{Methods } & \tabincell{c}{Time(per face)} \\
    \hline\hline
    CDCN & 120~ms \\
    \hline
    CFAN & 30~ms \\
    \hline
    TCDCN & 18~ms \\
    \hline
     Ours & 9~ms \\
    \hline
    \end{tabular}
    }
  \label{tab:landmark_detection_time}
\end{table}

\section{Conclusions}\label{sec:conclusion}
In this paper, we have presented a novel cascaded backbone-branches fully-convolutional network (BB-FCN) that progressively produces response maps of facial landmarks in an end-to-end manner. Our extensive experiments demonstrate that BB-FCN achieves very promising results on both traditional benchmarks with a controlled setting and on cluttered, real-world scenes. When exploiting our facial landmark localization results in R-CNN-based face detection, we have observed a significant increase in both accuracy and efficiency. In the future, we will integrate our BB-FCN model with object recognition and detection systems where accurate part-based localization can be helpful in improving object detection performance.


%





\ifCLASSOPTIONcaptionsoff
  \newpage
\fi



%

{
\small
\bibliography{reference}
\bibliographystyle{IEEEtran}
}

\vspace{-10mm}
\begin{IEEEbiography}[{\includegraphics[width=1in,height=1.25in,clip,keepaspectratio]{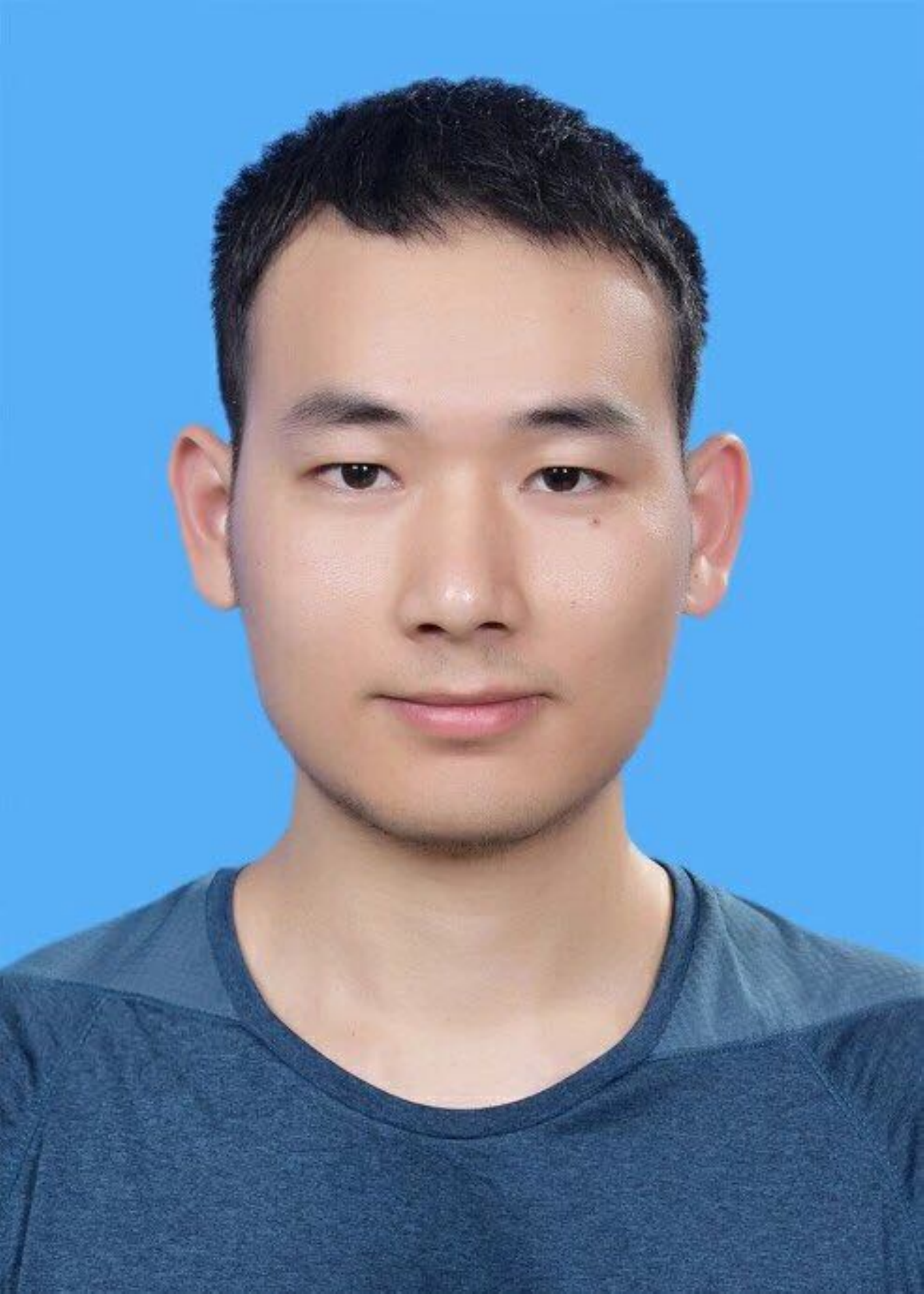}}]{Lingbo Liu} received his B.E. degree from the School of Software, Sun Yat-sen University, Guangzhou, China, in 2015, where he is currently pursuing his Ph.D. degree in computer science in the School of Data and Computer Science. His current research interests include computer vision, big data analysis, and intelligent transportation systems.
\end{IEEEbiography}

\vspace{-10mm}
\begin{IEEEbiography}[{\includegraphics[width=1in,height=1.25in,clip,keepaspectratio]{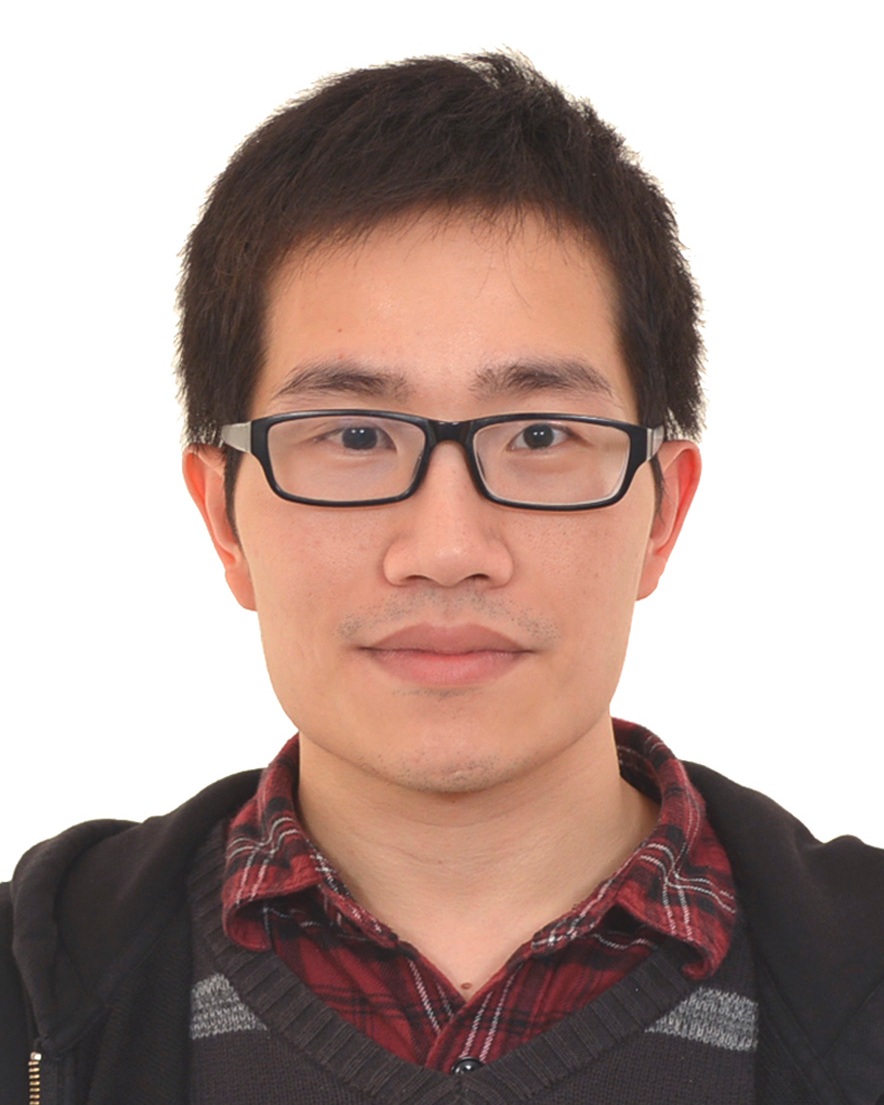}}]{Guanbin Li} is currently a research associate professor in the School of Data and Computer Science, Sun Yat-sen University. He received his PhD degree from the University of Hong Kong in 2016. He was a recipient of the Hong Kong Postgraduate Fellowship. His current research interests include computer vision, image processing, and deep learning. He has authored and co-authored more than 20 papers in top-tier academic journals and conferences. He serves as an area chair for the conference of VISAPP. He has been serving as a reviewer for  numerous academic journals and conferences, such as TPAMI, TIP, TMM, TC, CVPR2018 and IJCAI2018.
\end{IEEEbiography}

\vspace{-10mm}
\begin{IEEEbiography}[{\includegraphics[width=1in,height=1.25in,clip,keepaspectratio]{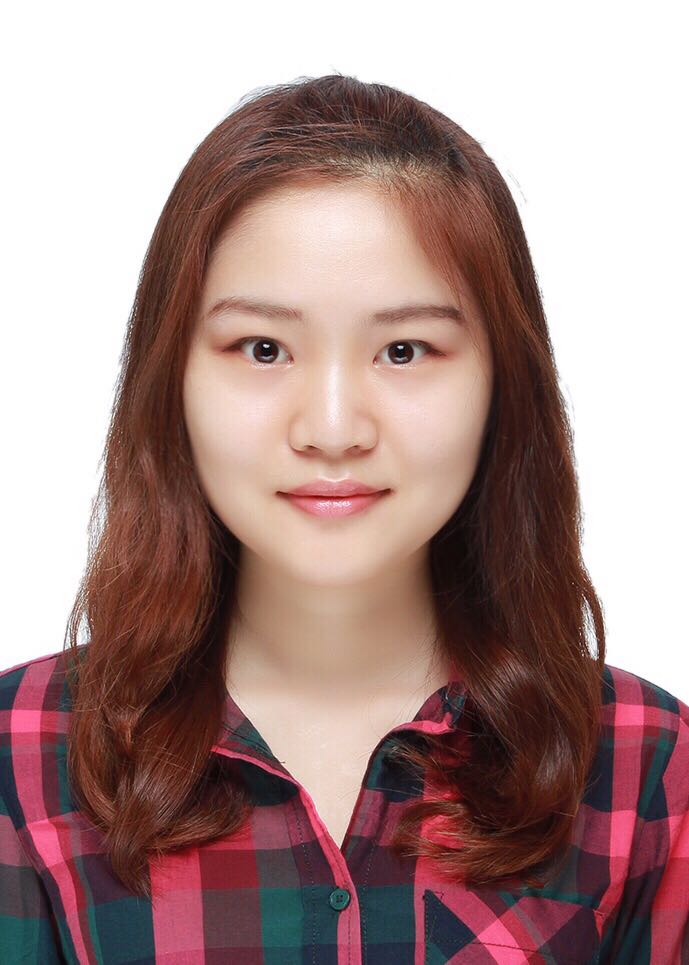}}]{Yuan Xie} received her BE degree and is currently working towards her ME degree in the School of Data and Computer Science, Sun Yat-sen University, China. Her research interests include facial alignment and salient object detection.
\end{IEEEbiography}

\vspace{-10mm}
\begin{IEEEbiography}[{\includegraphics[width=1in,height=1.25in,clip,keepaspectratio]{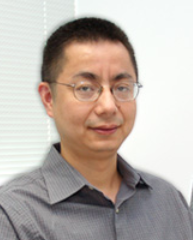}}]{Yizhou Yu} received his PhD degree from the University of California at Berkeley in 2000. He is currently a professor at The University of Hong Kong, and he was a faculty member at the University of Illinois, Urbana-Champaign, between 2000 and 2012. He is a recipient of the 2002 US National Science Foundation CAREER Award and 2007 NNSF China Overseas Distinguished Young Investigator Award. He has served on the editorial boards of IET Computer Vision, IEEE Transactions on Visualization and Computer Graphics, The Visual Computer, and International Journal of Software and Informatics. He has also served on the program committees of many leading international conferences, including SIGGRAPH, SIGGRAPH Asia, and International Conference on Computer Vision. His current research interests include deep learning methods for computer vision, computational visual media, geometric computing, video analytics and biomedical data analysis.
\end{IEEEbiography}

\vspace{-10mm}
\begin{IEEEbiography}[{\includegraphics[width=1in,height=1.25in,clip,keepaspectratio]{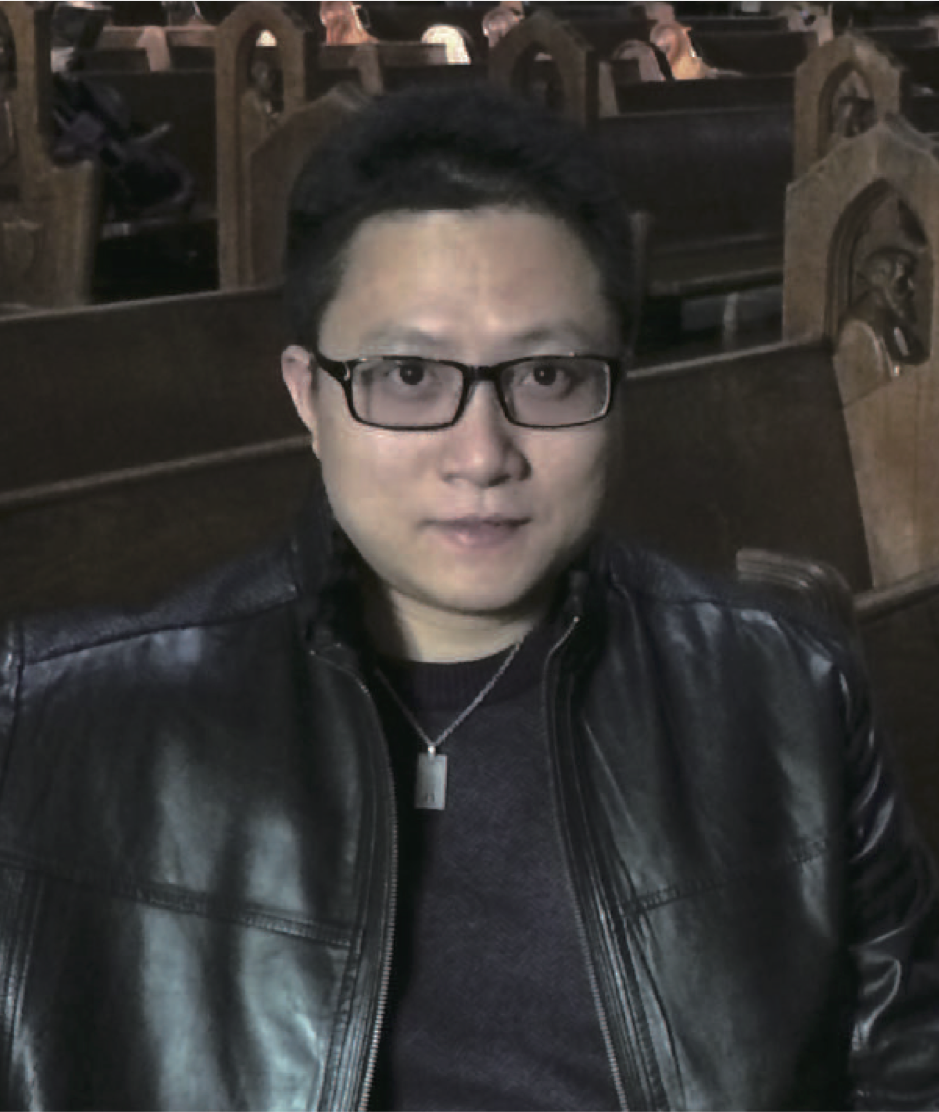}}]{Liang Lin}
(M'09, SM'15) is a full Professor of Sun Yat-sen University. He is an Excellent Young Scientist of the National Natural Science Foundation of China. From 2008 to 2010, he was a Post-Doctoral Fellow at the University of California, Los Angeles. From 2014 to 2015, as a senior visiting scholar, he was with The Hong Kong Polytechnic University and The Chinese University of Hong Kong. He currently leads the SenseTime R$\&$D teams to develop cutting-edge and deliverable solutions on computer vision, data analysis and mining, and intelligent robotic systems. He has authored and co-authored more than 100 papers in top-tier academic journals and conferences. He has been serving as an associate editor of IEEE Trans. Human-Machine Systems, The Visual Computer and Neurocomputing. He served as area/session chairs for numerous conferences, such as ICME, ACCV, ICMR. He was the recipient of the Best Paper Runners-Up Award in ACM NPAR 2010, the Google Faculty Award in 2012, the Best Paper Diamond Award in IEEE ICME 2017, and the Hong Kong Scholars Award in 2014. He is a Fellow of IET.
\end{IEEEbiography}




\end{document}